\documentclass{article}
\usepackage[main, final]{neurips_2026}
\usepackage[utf8]{inputenc}
\usepackage[T1]{fontenc}
\usepackage{url}
\usepackage{booktabs}
\usepackage{amsfonts}
\usepackage{amsmath}
\usepackage{amssymb}
\usepackage{nicefrac}
\usepackage{microtype}
\usepackage{xcolor}
\usepackage{graphicx}
\usepackage{mathtools}
\usepackage{amsthm}
\usepackage{multirow}
\usepackage{caption}
\usepackage[normalem]{ulem}
\useunder{\uline}{\ul}{}
\usepackage[breaklinks=true,colorlinks=true,linkcolor=blue,citecolor=blue,urlcolor=blue]{hyperref}

\newtheorem{proposition}{Proposition}

\newcommand{\method}{\textsc{SphereFlow}}

\title{\method{}: Missing Modality Imputation\\via Geometric Transport on Hypersphere}

\author{
  \textbf{Huihan Wang$^{1*}$\quad Juncheng Wang$^{1*}$\quad Yuxiang Feng$^{2,3}$\quad Wenlong Hou$^{1}$\quad Guoqi Yu$^{1}$} \\
  \textbf{Chao Xu$^{3,4}$\quad Yang Liu$^{3,4}$\quad Baigui Sun$^{3,4}$\quad Shujun Wang$^{1,5\dagger}$} \\[4pt]
  $^{1}$Department of Biomedical Engineering and Sports Technology (BEST), \\
  The Hong Kong Polytechnic University, Hong Kong, China \\
  $^{2}$Zhejiang University, Hangzhou, China \\
  $^{3}$IROOTECH TECHNOLOGY, China \\
  $^{4}$Wolf 1069 b Lab, Sany Group, China \\
  $^{5}$Department of Data Science and Artificial Intelligence, \\
  The Hong Kong Polytechnic University, Hong Kong, China \\[4pt]
  $^{*}$Equal contribution.\quad $^{\dagger}$Corresponding author. \\
  \texttt{\{wang2003huihan@gmail.com, wjc2830@gmail.com\}}$^{*}$,\quad \texttt{shu-jun.wang@polyu.edu.hk}$^{\dagger}$
}

\begin{document}

\maketitle

\begin{abstract}
Missing modality imputation aims to synthesize unobserved modalities from observed ones.
Existing methods treat each modality as living on its own latent manifold, forcing the generator to learn a long, unstructured jump between them.
We present \method{}, which takes a fundamentally different view: by projecting all modalities into a shared hyperspherical latent space via a frozen self-supervised encoder, cross-modal imputation reduces to \emph{short-range geometric transport} between nearby points on the sphere.
Realizing this idea with flow matching, however, exposes two geometric conflicts: the standard Gaussian noise source is catastrophically mismatched with the unit-norm data manifold, and the linear interpolation path departs the sphere surface into a region where the decoder has no training signal.
We resolve both with simple, geometry-aware modifications: replacing the noise source with the observed-modality latent so that the velocity field learns only the cross-modal displacement, and training the decoder with a time-weighted loss that extends its domain to the near-surface region traversed by the flow.
We prove that this data-as-source formulation reduces the worst-case off-manifold deviation from a quantity that grows with the latent dimension to a small, dimension-independent bound governed by inter-modality similarity.
Experiments on BraTS multi-modal MRI imputation and CT--MRI translation demonstrate state-of-the-art generation fidelity across all settings, while downstream brain tumor segmentation with our imputed data closes roughly eighty percent of the gap to the full-modality oracle. Code is available at \url{https://github.com/WTHH031230/SphereFlow}.
\end{abstract}

\section{Introduction}
\label{sec:intro}

\begin{figure}[t]
\centering
\includegraphics[width=\textwidth]{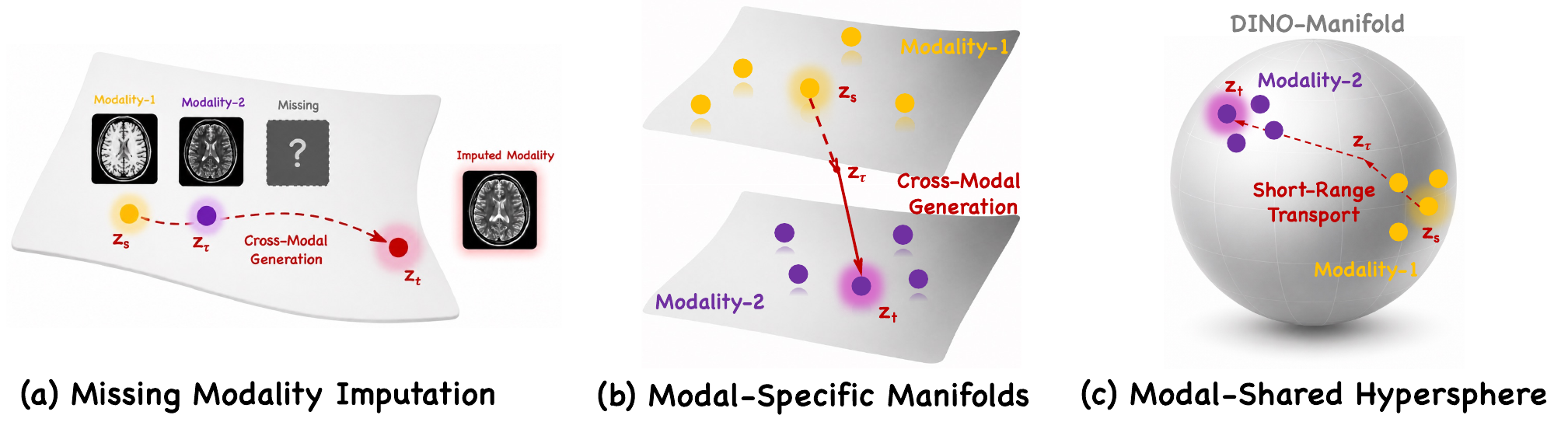}
\caption{\textbf{Two views of cross-modal imputation.} \textbf{(a)} Task: given a subset of observed modalities, synthesize the missing one. \textbf{(b)} Conventional view: modalities are encoded into disjoint latent manifolds (pixel space or per-modality VAE), so cross-modal generation is a long jump between them. \textbf{(c)} Our view: a frozen DINOv2 encoder maps every modality to a single shared hypersphere $\mathcal{S}^{d-1}$, on which paired latents $(z_s, z_t)$ sit close together; imputation becomes short-range transport along the sphere.}
\label{fig:teaser}
\end{figure}

Missing modality imputation---synthesizing an unobserved modality from observed ones---is critical in medical imaging~\citep{wang2021medical}, remote sensing, and autonomous driving, where acquiring all modalities is often infeasible~\citep{chartsias2018multimodal,sharma2019missing}. Existing approaches treat this as a conditional generation problem, leveraging generative adversarial networks~\citep{goodfellow2014gan,isola2017pix2pix,zhu2017cyclegan} or diffusion-based generative models~\citep{ho2020ddpm,song2021sde}, with designs ranging from modality-specific architectures~\citep{chartsias2018multimodal} and pairwise alignment losses~\citep{misaldm2025} to per-pair noise-to-data pipelines~\citep{dar2019image}. As illustrated in Fig.~\ref{fig:teaser}(b), all of these methods implicitly assume that each modality lives on its own latent manifold, so that imputation is a \emph{long, unstructured jump} between separated manifolds.

Nevertheless, these methods scale poorly with the number of modalities and, more importantly, neglect a fundamental principle: different modalities of the same subject share the same semantic content and differ only in appearance.
We argue that the bottleneck is \emph{representation}, not generation. Current methods operate in pixel space or VAE latent spaces~\citep{kingma2014auto,rombach2022high}, where an MRI and a CT scan of the same brain land in unrelated regions because their pixel statistics differ entirely---the model must learn a separate appearance-level mapping for every source-target pair. 

On the stark contrast, if all modalities were instead projected into a shared space organized by \emph{semantic content} (Fig.~\ref{fig:teaser}(c)), cross-modal imputation would reduce to \emph{transport}: a short-range displacement between nearby points. This raises a concrete question: \textbf{what properties must such a latent space satisfy so that missing modality imputation can be formulated as geometric transport?}

To answer this, we conduct pilot studies comparing three tokenizers---VAE~\citep{kingma2014auto}, MAE~\citep{he2022masked}, and DINOv2~\citep{caron2021dino,oquab2024dinov2}---as cross-modal latent spaces. Two properties emerge as essential: \textbf{(i)~cross-modal compactness relative to subject variability}---different modalities of the same subject should be encoded much closer than different subjects of the same modality---and \textbf{(ii)~geometric regularity}---latents should concentrate on a manifold with known geometry. VAE latents fail~(i): different modalities land far apart, as the encoder captures pixel statistics rather than semantics. MAE satisfies~(i) only by collapsing all subjects (regardless of modality) into a small region of latent space, sacrificing subject discriminability. DINOv2 uniquely satisfies both: its self-supervised contrastive training~\citep{chen2020simclr,he2020moco} yields features dominated by subject identity with small cross-modal displacement~(i), and its LayerNorm constrains representations to a hypersphere~$\mathcal{S}^{d-1}$~(ii)~\citep{dinosae2026,kumar2026manifold}.

Guided by these findings, we propose \method{}, which formulates cross-modal imputation as geometric transport on~$\mathcal{S}^{d-1}$. We adopt DINOv2 as a frozen encoder within a Representation Autoencoder (RAE)~\citep{zheng2025rae}, mapping all modalities into the same~$\mathcal{S}^{d-1}$ where cross-modal pairs of the same subject land nearby. Imputation becomes \textbf{short-range transport on~$\mathcal{S}^{d-1}$}: given a source-modality point~$z_s$, find the target-modality point~$z_t$.

However, realizing this with flow matching~\citep{lipman2023flow,liu2022rectified,esser2024sd3} exposes two geometric conflicts. \textbf{First}, standard flow matching generates a target latent~$z$ from Gaussian noise $\varepsilon \sim \mathcal{N}(0, I)$---a paradigm designed for unconditional generation from scratch, not for cross-modal translation where a semantically meaningful source is already available. On a hypersphere this mismatch is severe: $\|\varepsilon\|\!\approx\!\sqrt{d}$ while $\|z\|{=}1$, so the interpolant must traverse a vast off-manifold void and the model wastes capacity on a radial collapse irrelevant to cross-modal transformation. \textbf{Second}, even with both endpoints on the sphere, linear interpolation produces intermediates with~$\|z_\tau\|<1$---off-manifold points the decoder has never seen and cannot reconstruct. Riemannian flow matching~\citep{chen2023riemannian,kumar2026manifold} fixes the path via SLERP but still departs from noise; LV-RAE~\citep{lvrae2026} augments the decoder but without geometric structure. No existing method addresses both.

We present \method{} to address both conflicts. For the first, we replace the Gaussian source with the paired source-modality feature---already on~$\mathcal{S}^{d-1}$---so the velocity field learns only the modality-specific displacement. For the second, we train the decoder to reconstruct from any interpolant~$z_\tau$ with time-dependent weights~$(1{-}\tau)$ and~$\tau$, extending its domain to the near-surface region the flow traverses. We validate on BraTS~\citep{brats2021} (T1, T1ce, T2, FLAIR) and a paired CT--MRI benchmark, achieving state-of-the-art generation fidelity across all settings. The imputed modalities also improve downstream performance---brain tumor segmentation with our synthesized data approaches the accuracy of training with real complete data.

Our contributions:
\begin{itemize}
    \item We formulate missing modality imputation as \emph{geometric transport on a hypersphere}, identifying two essential latent space properties through pilot studies and showing that DINOv2-based RAE uniquely satisfies both.
    \item We identify two geometric conflicts when applying flow matching on~$\mathcal{S}^{d-1}$ and address both: replacing the noise source with the observed modality, and expanding the decoder's domain with time-weighted reconstruction.
    \item We show that data-as-source reduces off-manifold deviation from~$\Theta(\sqrt{d}\,)$ to a small bounded quantity controlled by inter-modality cosine similarity.
    \item We achieve state-of-the-art results on BraTS and CT--MRI translation in both generation fidelity and downstream task performance.
\end{itemize}

\section{Pilot Study: What Makes a Good Shared Latent Space?}
\label{sec:pilot}

\begin{figure}[t]
\centering
\includegraphics[width=\textwidth]{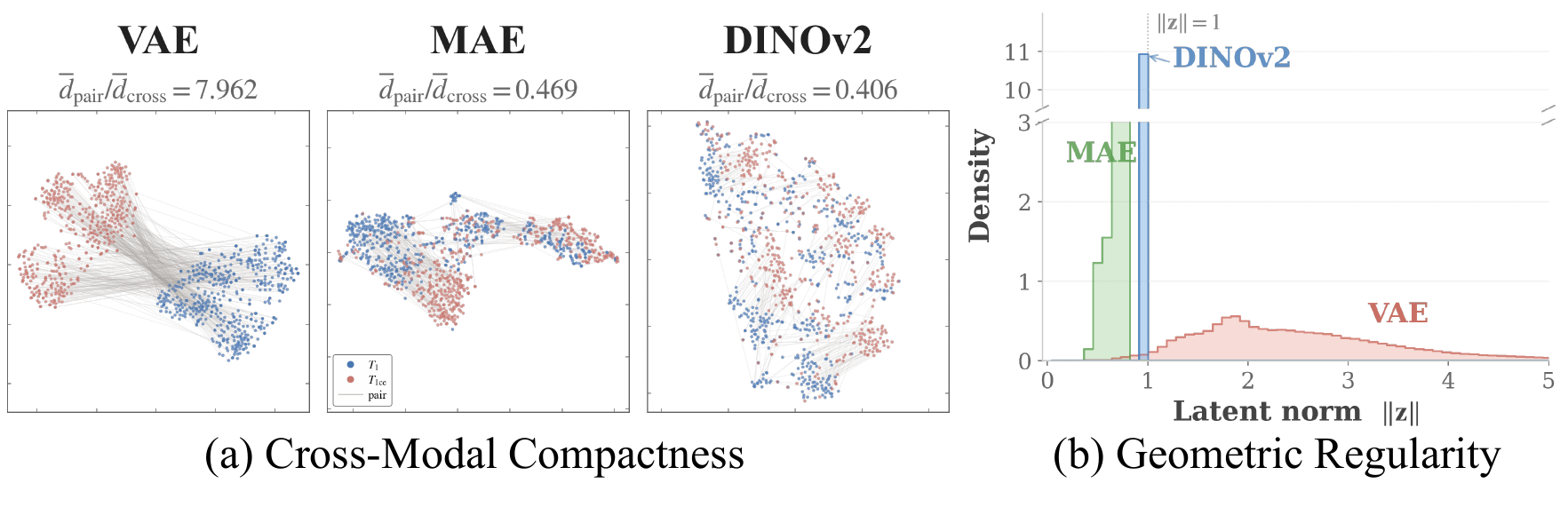}
\caption{\textbf{Pilot study: tokenizer comparison on BraTS.} \textbf{(a)} Cross-modal compactness: 2-D projection of paired $T_1$ (blue) and $T_{1c}$ (red) latents under three tokenizers, with thin lines connecting same-subject pairs; the title reports the ratio $\bar d_{\text{pair}}/\bar d_{\text{cross}}$ (lower~is~better, see~\S\ref{sec:pilot_compact}). \textbf{(b)} Geometric regularity: empirical distribution of per-token latent norms $\|z\|$.}
\label{fig:pilot}
\end{figure}

In Section~\ref{sec:intro}, we argued that transport-based imputation requires two latent space properties: \emph{cross-modal compactness} (paired modalities closer than different subjects) and \emph{geometric regularity} (concentration on a known manifold). We now empirically test these on three representative tokenizers---VAE~\citep{kingma2014auto}, MAE~\citep{he2022masked}, and DINOv2~\citep{oquab2024dinov2}---by encoding paired $T_1$/$T_{1c}$ images from BraTS~\citep{brats2021} (Fig.~\ref{fig:pilot}).

\subsection{Cross-Modal Compactness}
\label{sec:pilot_compact}

\paragraph{Setup.}
We extract paired $T_1$/$T_{1c}$ latents from 500 BraTS test slices (L2-normalized) and compute two cosine-distance statistics: $d_{\text{cross}}$, the mean distance between modalities of the \emph{same} slice, and $d_{\text{intra}}$, the mean distance between \emph{different} slices within the same modality. A good transport space needs a small ratio $d_{\text{cross}}/d_{\text{intra}}$ without $d_{\text{intra}}$ itself collapsing (which would indicate loss of anatomical discriminability).

\paragraph{Findings.}
As shown in Fig.~\ref{fig:pilot}(a), VAE gives $d_{\text{cross}}/d_{\text{intra}}\!\approx\!7.96$: modalities land in unrelated regions, confirming that pixel-level encoders force long, unstructured jumps. MAE reduces the ratio to $0.47$ but via representational collapse---all latents cluster into a tiny region, erasing anatomical identity. DINOv2 achieves the lowest ratio ($0.41$) while preserving well-separated slices: cross-modal displacement is small \emph{relative to} anatomical variability. \textbf{Takeaway:} only DINOv2 provides cross-modal compactness without sacrificing discriminability, making short-range transport viable.

\subsection{Geometric Regularity}
\label{sec:pilot_regular}

\paragraph{Setup.}
We examine the distribution of latent norms~$\|z\|$ for each tokenizer and compute the coefficient of variation $\text{CV}_r = \text{std}(\|z\|)/\mathbb{E}[\|z\|]$. A small~$\text{CV}_r$ indicates concentration on a hypersphere, enabling geometric analysis of interpolation paths and decoder domains.

\paragraph{Findings.}
Figure~\ref{fig:pilot}(b) shows the results. VAE latents spread broadly across~$\mathbb{R}^d$ with no manifold structure. MAE latents show moderate radial concentration but lack a clean single-radius manifold. DINOv2 latents concentrate sharply at~$\|z\|\!\approx\!1$ with near-zero~$\text{CV}_r$, consistent with its output LayerNorm---confirming that they lie on a hypersphere~$\mathcal{S}^{d-1}$.
\textbf{Takeaway:} DINOv2's hyperspherical geometry provides a well-characterized manifold for analyzing flow paths, decoder domains, and transport bounds.

\paragraph{Summary.}
Among the three tokenizers, only DINOv2 simultaneously satisfies both properties: cross-modal compactness without collapse and concentration on a known manifold~$\mathcal{S}^{d-1}$. We therefore adopt DINOv2 as a frozen encoder within a Representation Autoencoder (RAE), formulating imputation as short-range transport on the hypersphere. However, applying flow matching on~$\mathcal{S}^{d-1}$ introduces geometric conflicts that we analyze and resolve in Section~\ref{sec:method}.

\section{Method: \method{}}
\label{sec:method}

\begin{figure}[t]
\centering
\includegraphics[width=\textwidth]{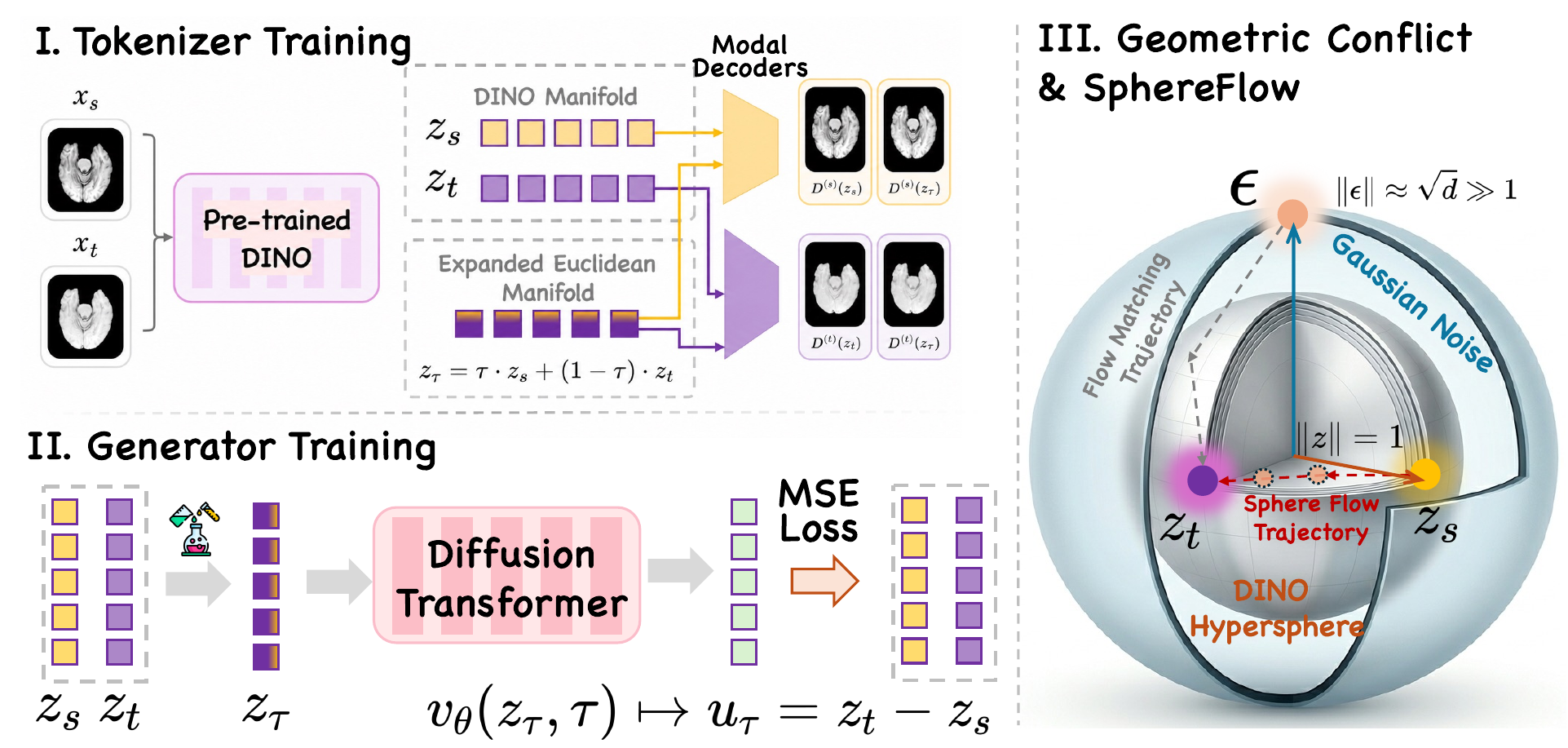}
\caption{\textbf{Overview of \method{}.} \textbf{(I)~Tokenizer training} (\S\ref{sec:decoder_expand}): a frozen DINOv2 encoder maps source and target images onto the hypersphere~$\mathcal{S}^{d-1}$. Modality-specific decoders~$D^{(s)}\!,D^{(t)}$ are trained to reconstruct both on-manifold latents ($z_s, z_t$) and off-manifold interpolants ($z_\tau$) via the time-weighted expansion loss (Eq.~\ref{eq:expand_loss}). \textbf{(II)~Generator training} (\S\ref{sec:cmfm}): a Diffusion Transformer learns the cross-modal velocity field $v_\theta(z_\tau, \tau) \mapsto u_\tau = z_t - z_s$ with $z_\tau$ as input, using the paired source latent as the flow origin instead of Gaussian noise. \textbf{(III)~Geometric conflict \& \method{}}: standard flow matching (gray trajectory) departs from Gaussian noise at $\|\varepsilon\|\!\approx\!\sqrt{d}\gg 1$ and must traverse a vast off-manifold void to reach the DINO hypersphere ($\|z\|{=}1$). \method{} (orange trajectory) transports directly from~$z_s$ to~$z_t$ along a short near-surface chord, with bounded off-manifold deviation (Proposition~\ref{prop:bound}).}
\label{fig:method}
\end{figure}

The pilot study establishes DINOv2 as a shared hyperspherical latent space where cross-modal imputation reduces to short-range transport on~$\mathcal{S}^{d-1}$. However, na\"ively applying flow matching to this setting introduces two geometric conflicts that must be resolved. This section presents \method{}. We begin with the baseline pipeline (\S\ref{sec:prelim}), formalize these conflicts (\S\ref{sec:conflicts}), and then introduce our solutions: cross-modal flow matching (\S\ref{sec:cmfm}) and manifold-expanded decoder training (\S\ref{sec:decoder_expand}). Figure~\ref{fig:method} provides a high-level overview.

\subsection{Preliminaries}
\label{sec:prelim}

\paragraph{DINO Encoder.}
We adopt a frozen DINOv2 ViT-B/14~\citep{oquab2024dinov2} as our visual tokenizer. Given an input image~$x \in \mathbb{R}^{H \times W \times C}$, we denote its output representation as $z = \mathcal{E}_{\text{DINO}}(x)$. Due to the final LayerNorm, DINOv2 representations satisfy $\|z\| \approx 1$ with near-zero radial variance~\citep{dinosae2026,kumar2026manifold}, so we treat $z \in \mathcal{S}^{d-1}$ throughout. The encoder is \emph{frozen} and shared across all modalities, guaranteeing a common geometric space.

\paragraph{RAE Decoder.}
Following~\citet{zheng2025rae}, we pair the frozen DINO encoder with \emph{modality-specific} decoders~$\{D_\phi^{(m)}\}_{m=1}^M$, one per modality, each reconstructing images of its respective modality from latent representations. Each decoder takes~$z \in \mathcal{S}^{d-1}$ as input and produces~$\hat{x} = D_\phi^{(m)}(z) \in \mathbb{R}^{H \times W \times C}$. They are trained with pixel-level and perceptual losses:
\begin{equation}
    \mathcal{L}_{\text{rec}} = \| D_\phi^{(m)}(z) - x \|_1
            + \lambda_{\text{perc}} \cdot \mathcal{L}_{\text{LPIPS}}(D_\phi^{(m)}(z), x)
            + \lambda_{\text{adv}} \cdot \mathcal{L}_{\text{adv}}(D_\phi^{(m)}(z))
    \label{eq:rec_loss}
\end{equation}
where $\mathcal{L}_{\text{LPIPS}}$ denotes the perceptual loss and $\mathcal{L}_{\text{adv}}$ a non-saturating GAN loss against a hinge-based discriminator. Using modality-specific decoders allows each to specialize in the pixel statistics of its modality (e.g., intensity ranges of T1 vs.\ T2), while sharing the same latent space. A critical observation for our later analysis: each decoder is trained \emph{exclusively} on on-manifold points~$z = \mathcal{E}_{\text{DINO}}(x) \in \mathcal{S}^{d-1}$. Its behavior at off-manifold points ($\|z\| \neq 1$) is uncontrolled and, as we will show, problematic.

\paragraph{Diffusion Transformer (DiT).}
We employ a Diffusion Transformer (DiT)~\citep{peebles2023dit}, which adapts the Vision Transformer~\citep{vaswani2017attention,dosovitskiy2021vit} for diffusion modeling, as the velocity field backbone. The DiT takes a noisy latent~$z_\tau$ at timestep~$\tau \in [0,1]$ and predicts the velocity~$v_\theta(z_\tau, \tau, c)$, where $c$ is a modality indicator specifying the target modality. This enables a single model to handle all imputation directions.

\paragraph{Standard Flow Matching.}
Flow matching~\citep{lipman2023flow} learns a velocity field that defines a continuous-time ordinary differential equation (ODE) transporting a source distribution~$p_0$ to a target distribution~$p_1$. In the standard formulation, the source is Gaussian noise~$\varepsilon \sim \mathcal{N}(0, I)$ and the target is data~$z = \mathcal{E}_{\text{DINO}}(x)$. The conditional interpolant and velocity at timestep~$\tau$ are:
\begin{equation}
    z_\tau = \tau \cdot z + (1{-}\tau) \cdot \varepsilon, \quad u_\tau = z - \varepsilon
    \label{eq:standard_fm}
\end{equation}
The DiT is trained to predict the velocity via:
\begin{equation}
    \mathcal{L}_{\text{FM}} = \mathbb{E}_{\tau \sim \mathcal{U}(0,1),\, \varepsilon \sim \mathcal{N}(0,I),\, z} \left\| v_\theta(z_\tau, \tau) - u_\tau \right\|^2
    \label{eq:fm_loss}
\end{equation}
At inference, a sample~$\varepsilon$ is drawn and integrated forward via the learned ODE to obtain~$\hat{z} \approx z$, which is then decoded as~$\hat{x} = D_\phi(\hat{z})$.

\subsection{Geometric Conflicts on $\mathcal{S}^{d-1}$}
\label{sec:conflicts}

With the shared hyperspherical latent space established, a natural next step is to apply flow matching to learn the cross-modal transport map on~$\mathcal{S}^{d-1}$. However, directly applying the standard formulation (Eq.~\ref{eq:standard_fm}) to this geometry introduces two fundamental conflicts.

\paragraph{Conflict I: Source Distribution Mismatch.}
Data lies on~$\mathcal{S}^{d-1}$ ($\|z\| = 1$), while the Gaussian source concentrates at~$\|\varepsilon\| \approx \sqrt{d}$ (e.g., ${\approx}27.7$ for $d{=}768$). The interpolant~$z_\tau$ (Eq.~\ref{eq:standard_fm}) therefore traverses norms from~$\sqrt{d}$ to~$1$, and the velocity field must simultaneously learn (a)~a radial contraction by~$\sqrt{d}$, (b)~a directional rotation, and (c)~the modality-specific displacement---three entangled tasks, only the last of which is relevant. In cross-modal imputation this is especially wasteful: the observed modality's latent~$z_s \in \mathcal{S}^{d-1}$ is already a semantically meaningful source, yet the standard formulation discards it in favor of noise.

\paragraph{Conflict II: Off-Manifold Decoder Blindness.}
More broadly, flow matching with linear interpolation cannot guarantee that the entire trajectory stays on~$\mathcal{S}^{d-1}$. For any two distinct points $z_0, z_1 \in \mathcal{S}^{d-1}$, the linear interpolant $z_\tau = \tau \cdot z_1 + (1{-}\tau) \cdot z_0$ satisfies
\begin{equation}
    \|z_\tau\|^2 = \tau^2 + (1{-}\tau)^2 + 2\tau(1{-}\tau)\langle z_0, z_1 \rangle < 1 \quad \text{whenever } \langle z_0, z_1 \rangle < 1
    \label{eq:norm_deficit}
\end{equation}
That is, intermediate states inevitably dip \emph{inside} the sphere. However, each RAE decoder~$D_\phi^{(m)}$ is trained exclusively on on-manifold points ($\|z\| = 1$) and has never seen inputs with $\|z\| < 1$. During inference, the flow trajectory passes through this unseen region, and the decoders cannot provide meaningful reconstructions there---a mismatch between where the flow operates and where the decoders are valid.

\subsection{Cross-Modal Flow Matching}
\label{sec:cmfm}

To resolve Conflict~I, we replace the Gaussian source with the paired source-modality latent. Let $z_s = \mathcal{E}_{\text{DINO}}(x_s) \in \mathcal{S}^{d-1}$ and $z_t = \mathcal{E}_{\text{DINO}}(x_t) \in \mathcal{S}^{d-1}$ be a cross-modal pair. The interpolant and velocity are:
\begin{equation}
    z_\tau = \tau \cdot z_t + (1{-}\tau) \cdot z_s, \quad u_\tau = z_t - z_s
    \label{eq:cmfm}
\end{equation}
The velocity~$u_\tau$ is constant along the path---it is simply the displacement from source to target latent. The DiT is trained to predict this displacement conditioned on the target modality indicator~$c$:
\begin{equation}
    \mathcal{L}_{\text{CMFM}} = \mathbb{E}_{\tau \sim \mathcal{U}(0,1),\, (z_s, z_t)} \left\| v_\theta(z_\tau, \tau, c) - (z_t - z_s) \right\|^2
    \label{eq:cmfm_loss}
\end{equation}

Both endpoints now lie on~$\mathcal{S}^{d-1}$, eliminating the norm mismatch, and the velocity field learns only the modality-specific displacement. Moreover, the interpolant remains close to the manifold, as quantified by the following result.

\begin{proposition}[Off-manifold deviation bound]
\label{prop:bound}
Let $z_s, z_t \in \mathcal{S}^{d-1}$ with $\langle z_s, z_t \rangle = \cos\alpha$ for angular separation~$\alpha$. The maximum off-manifold deviation of the linear interpolant~$z_\tau$ (Eq.~\ref{eq:cmfm}) is:
\begin{equation}
    \Delta \;\coloneqq\; \max_{\tau \in [0,1]} \left(1 - \|z_\tau\|\right) = 1 - \sqrt{\frac{1 + \cos\alpha}{2}} = 1 - \cos\frac{\alpha}{2}
    \label{eq:deviation_bound}
\end{equation}
attained at $\tau = \tfrac{1}{2}$. By contrast, for Gaussian-source flow matching, $\mathbb{E}\bigl[\max_\tau \bigl|\,\|z_\tau\| - 1\,\bigr|\bigr] = \Theta(\sqrt{d}\,)$.
\end{proposition}

The proof is given in Appendix~\ref{app:proof_bound}.

Note that $\Delta$ depends only on the angular separation~$\alpha$ between the source and target latents, not on the ambient dimension~$d$. In missing modality imputation, the two modalities depict the same underlying anatomy or scene, so their DINO representations naturally share substantial semantic overlap---$\alpha$ is small by the nature of the task. This contrasts sharply with the Gaussian-source deviation~$\Theta(\sqrt{d}\,)$, which grows with dimension. The bounded, dimension-independent deviation is what makes the decoder expansion in~\S\ref{sec:decoder_expand} tractable.

At inference, given a source latent~$z_s = \mathcal{E}_{\text{DINO}}(x_s)$ and a target modality indicator~$c$, we integrate the learned velocity field starting from~$z_s$:
\begin{equation}
    \hat{z}_t = z_s + \int_0^1 v_\theta(z_\tau, \tau, c)\, d\tau
\end{equation}
and decode with the target-modality decoder:~$\hat{x}_t = D_\phi^{(t)}(\hat{z}_t)$.

\subsection{Manifold-Expanded Decoder Training}
\label{sec:decoder_expand}

Cross-modal flow matching (\S\ref{sec:cmfm}) resolves the source distribution mismatch (Conflict~I), but Conflict~II remains: the linear interpolant $z_\tau$ passes through the interior of the sphere, where the decoders have no training signal. We address this by expanding each decoder's domain to cover the off-manifold region that the flow traverses.

\paragraph{Time-Weighted Reconstruction Loss.}
We train each modality-specific decoder to reconstruct its corresponding image from any intermediate point~$z_\tau$ along the interpolation path, with time-dependent weights:
\begin{equation}
    \mathcal{L}_{\text{expand}} = \mathbb{E}_{\tau \sim \mathcal{U}(0,1)} \left[ \tau \cdot \ell\!\left(D_\phi^{(t)}(z_\tau),\, x_t\right) + (1{-}\tau) \cdot \ell\!\left(D_\phi^{(s)}(z_\tau),\, x_s\right) \right]
    \label{eq:expand_loss}
\end{equation}
where $\ell(\cdot, \cdot)$ denotes the reconstruction loss (Eq.~\ref{eq:rec_loss}) and $z_\tau = \tau \cdot z_t + (1{-}\tau) \cdot z_s$.

The $\tau$-weighting is principled: at $\tau{=}0$ each decoder reconstructs its own modality (weight~$1$); at $\tau{=}1$ the target decoder bears full responsibility. This matches supervision to the latent blend, avoiding conflicting gradients from a mismatched weighting.

\paragraph{Training Pipeline.}
\method{} is trained in two independent stages:

\emph{Stage~1: Tokenizer training.} We first train all modality-specific decoders~$\{D_\phi^{(m)}\}$ with the DINO encoder frozen, using paired multi-modal data. Each decoder is trained with both the standard on-manifold reconstruction loss (Eq.~\ref{eq:rec_loss}) for its own modality, and the manifold-expanded loss over cross-modal pairs $(x_s, x_t)$:
\begin{equation}
    \mathcal{L}_{\text{tokenizer}} = \mathcal{L}_{\text{rec}} + \lambda \cdot \mathcal{L}_{\text{expand}}
    \label{eq:tokenizer_loss}
\end{equation}
where $\lambda$ balances the two terms. The $\mathcal{L}_{\text{rec}}$ term ensures faithful on-manifold reconstruction for each modality, while $\mathcal{L}_{\text{expand}}$ extends all decoders' domains into the off-manifold region that the flow will later traverse.
Finally, since DINOv2 is a general-purpose encoder, we adapt it to the medical imaging modalities: we unfreeze the shared encoder~$\mathcal{E}_\psi$ (initialized from DINOv2) and train it jointly with all decoders, adding an alignment loss to a frozen copy~$\mathcal{E}_0$ of the pretrained encoder:
\begin{equation}
    \mathcal{L}_{\text{align}} = \lambda_{\cos}\bigl(1 - \cos(z, z')\bigr) + \lambda_{\text{mse}} \bigl\|\mathrm{GAP}(z) - \mathrm{GAP}(z')\bigr\|_2^2, \quad z = \mathcal{E}_\psi(x),\ z' = \mathrm{sg}\bigl(\mathcal{E}_0(x)\bigr)
    \label{eq:align_loss}
\end{equation}
where $\cos(\cdot,\cdot)$ is the token-level cosine similarity, $\mathrm{GAP}$ is global average pooling, and $\mathrm{sg}(\cdot)$ is the stop-gradient. The first term anchors token directions and the second anchors global semantics, letting the encoder adapt to the reconstruction and expansion objectives while staying close to the pretrained representation. After this stage, the tokenizer (aligned encoder + trained decoders) is fixed.

\emph{Stage~2: Flow matching training.} With the tokenizer frozen, we train the DiT independently with~$\mathcal{L}_{\text{CMFM}}$ (Eq.~\ref{eq:cmfm_loss}). Since the decoders were already expanded in Stage~1, they can reconstruct from any point along the flow trajectory at inference without further fine-tuning.

This decoupled design has a practical advantage: the tokenizer (encoder + all modality-specific decoders) is trained once, while the DiT can be trained or fine-tuned for specific modality pairs without retraining any decoder.

\section{Experiments}
\label{sec:experiments}
\vspace{-2pt}

\begin{figure}[t]
\centering
\includegraphics[width=\textwidth]{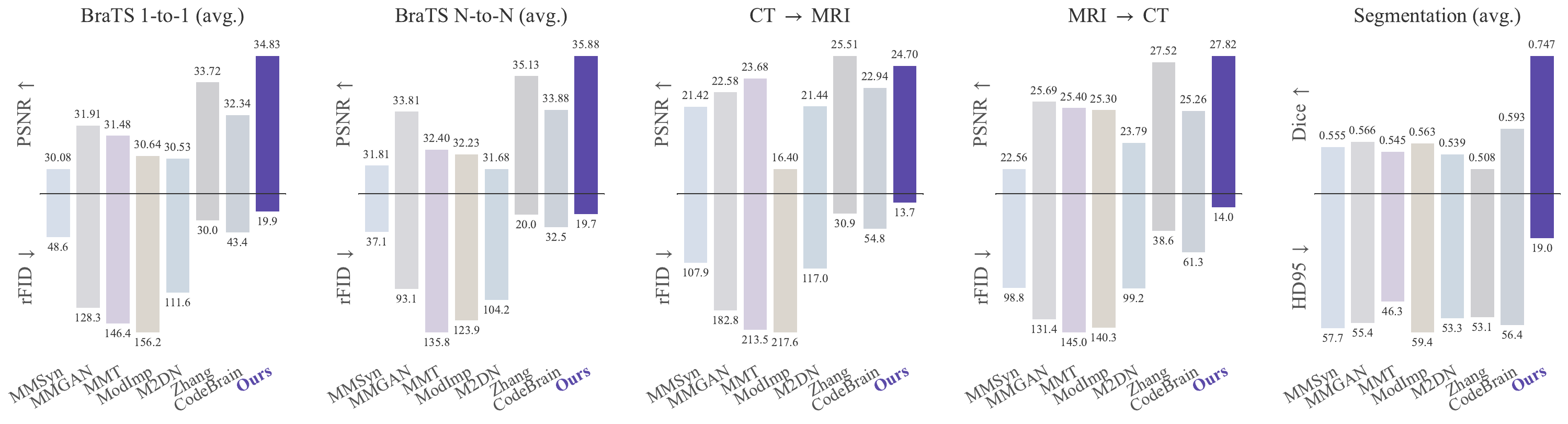}
\caption{\textbf{Main-results overview.} Top: PSNR/Dice ($\uparrow$); bottom: rFID/HD95 ($\downarrow$) across four generation tasks and segmentation. \method{} achieves the best or near-best score on every panel.}
\label{fig:main_tradeoff}
\end{figure}

We evaluate \method{} on multi-contrast MRI synthesis (BraTS,~\S\ref{sec:exp_brats}), CT--MRI translation (SynthRAD2023,~\S\ref{sec:exp_ctmri}), downstream segmentation under missing modalities (\S\ref{sec:exp_seg}), and ablations (\S\ref{sec:exp_ablation}). Figure~\ref{fig:main_tradeoff} previews the headline results.

\subsection{Experimental Setup}
\label{sec:exp_setup}

\paragraph{Datasets.}
\textbf{BraTS}~\citep{brats2021} provides four paired brain MRI modalities ($T_1$, $T_{1c}$, $T_2$, FLAIR/$T_f$); \textbf{SynthRAD2023}~(brain track)~\citep{synthrad2023} provides paired CT and $T_1$-weighted MRI. We extract axial $256{\times}256$ slices after skull-stripping and per-volume normalization, and follow official splits.

\paragraph{Implementation.}
We use frozen \textbf{DINOv2-B/14}~\citep{oquab2024dinov2} with modality-specific ViT-G/14 RAE decoders~\citep{zheng2025rae}. Stage~1 trains decoders in three steps: single-modality warm-up (Eq.~\ref{eq:rec_loss}), manifold-expansion fine-tuning (Eq.~\ref{eq:expand_loss}), and shared-encoder alignment. Stage~2 trains a \textbf{DiTDH-XL}~\citep{peebles2023dit} with the cross-modal flow loss (Eq.~\ref{eq:cmfm_loss}) and logit-normal time sampling for up to 400~epochs. Inference uses $2$-step Euler ODE on BraTS and $4$-step on SynthRAD. For N-to-N, available source latents are averaged into a single $z_s$.

\paragraph{Baselines and Metrics.}
We compare seven baselines: \textbf{MMSyn}~\citep{chartsias2018multimodal}, \textbf{MMGAN}~\citep{sharma2019missing}, \textbf{MMT}~\citep{liu2023mmt}, \textbf{M2DN}~\citep{meng2024m2dn}, \textbf{ModImp}~\citep{qi2025modimp}, \textbf{CodeBrain}~\citep{wu2025codebrain}, and \textbf{Zhang}~\citep{zhang2024unified}, all trained on the same splits. We report PSNR, SSIM~\citep{wang2004ssim,hore2010psnr}, paired rFID~\citep{heusel2017fid}, IS~\citep{salimans2016is}, and iFID~\citep{xu2026ifid} for generation quality, plus Dice, HD95, IoU, Sensitivity for downstream segmentation (3D~U-Net trained on full-modality BraTS and evaluated on imputed inputs).

\vspace{-2pt}
\subsection{Multi-Modal MRI Imputation on BraTS}
\label{sec:exp_brats}
\vspace{-2pt}

\paragraph{1-to-1 Translation.}
Table~\ref{tab:121_metrics_brats} reports all $12$~directions among $\{T_1, T_{1c}, T_2, T_f\}$. \method{} achieves the best PSNR on every direction (avg $34.83$\,dB, $+1.11$\,dB over runner-up) and reduces rFID by $33.6\%$ ($29.99{\to}19.93$), with the largest gains on hard pairs ($T_2{\to}T_{1c}$: $+1.24$\,dB).

\begin{table}[t]
    \centering
    \caption{1-to-1 generation on BraTS (all 12 directions).}
    \label{tab:121_metrics_brats}
    \resizebox{\textwidth}{!}{
    \begin{tabular}{llccccccccccccccccccc}
  \multicolumn{1}{c}{\multirow{2}{*}{Modalities}} & \multicolumn{1}{c}{} & \multicolumn{8}{c}{PSNR ↑}                                                                              & \multicolumn{1}{l}{} & \multicolumn{8}{c}{SSIM ↑}                                                                                                  \\
  \multicolumn{1}{c}{}                            & \multicolumn{1}{c}{} & MMSyn       & MMGAN  & MMT    & ModImp & M2DN   & Zhang          & CodeBrain   & \textbf{Ours} & \multicolumn{1}{l}{} & MMSyn         & MMGAN      & MMT                  & ModImp & M2DN   & Zhang         & CodeBrain     & \textbf{Ours} \\ \hline
  $T_1 \rightarrow T_{1c}$                        &                      & 30.50       & 29.93  & 30.75  & 31.38  & 29.84  & 32.79          & {\ul 32.85} & \textbf{33.53}      &                      & 0.9185        & 0.9075     & 0.9077               & 0.9235 & 0.8840 & 0.9384        & {\ul 0.9405}  & \textbf{0.9414}     \\
  $T_1 \rightarrow T_2$                           &                      & 27.91       & 29.68  & 29.62  & 29.29  & 28.87  & {\ul 31.76}    & 30.77       & \textbf{32.68}      &                      & 0.8985        & 0.9035     & 0.8916               & 0.9010 & 0.8675 & {\ul 0.9326}  & 0.9289        & \textbf{0.9390}     \\
  $T_1 \rightarrow T_f$                           &                      & 33.16       & 34.92  & 34.72  & 33.81  & 32.88  & {\ul 36.06}    & 35.36       & \textbf{37.35}      &                      & 0.9220        & 0.9317     & 0.9244               & 0.9271 & 0.8937 & {\ul 0.9470}  & 0.9458        & \textbf{0.9529}     \\
  $T_{1c} \rightarrow T_1$                        &                      & 34.82       & 36.81  & 33.65  & 33.83  & 33.50  & {\ul 37.21}    & 36.34       & \textbf{37.87}      &                      & 0.9421        & 0.9517     & 0.9235               & 0.9392 & 0.9067 & {\ul 0.9586}  & 0.9575        & \textbf{0.9612}     \\
  $T_{1c} \rightarrow T_2$                        &                      & 28.86       & 31.15  & 29.45  & 28.80  & 29.05  & {\ul 31.80}    & 30.58       & \textbf{32.67}      &                      & 0.9019        & 0.9183     & 0.8872               & 0.8952 & 0.8683 & {\ul 0.9305}  & 0.9252        & \textbf{0.9379}     \\
  $T_{1c} \rightarrow T_f$                        &                      & 33.19       & 34.90  & 33.64  & 33.67  & 33.43  & {\ul 35.60}    & 35.24       & \textbf{37.49}      &                      & 0.9147        & 0.9326     & 0.9148               & 0.9242 & 0.8972 & {\ul 0.9472}  & 0.9463        & \textbf{0.9535}     \\
  $T_2 \rightarrow T_1$                           &                      & 30.57       & 32.42  & 31.82  & 28.86  & 30.49  & {\ul 34.64}    & 31.72       & \textbf{35.63}      &                      & 0.9200        & 0.9251     & 0.9108               & 0.9097 & 0.8874 & {\ul 0.9509}  & 0.9324        & \textbf{0.9562}     \\
  $T_2 \rightarrow T_{1c}$                        &                      & 26.06       & 26.51  & 28.57  & 26.01  & 26.80  & {\ul 30.81}    & 27.50       & \textbf{32.05}      &                      & 0.8749        & 0.8908     & 0.8802               & 0.8795 & 0.8610 & {\ul 0.9260}  & 0.9031        & \textbf{0.9349}     \\
  $T_2 \rightarrow T_f$                           &                      & 30.48       & 35.20  & 34.25  & 31.87  & 33.02  & {\ul 37.15}    & 33.57       & \textbf{38.11}      &                      & 0.9060        & 0.9323     & 0.9146               & 0.9047 & 0.8891 & {\ul 0.9555}  & 0.9312        & \textbf{0.9584}     \\
  $T_f \rightarrow T_1$                           &                      & 31.34       & 34.08  & 32.52  & 31.80  & 31.12  & {\ul 34.28}    & 33.13       & \textbf{35.80}      &                      & 0.9243        & 0.9349     & 0.9174               & 0.9226 & 0.8931 & {\ul 0.9483}  & 0.9438        & \textbf{0.9548}     \\
  $T_f \rightarrow T_{1c}$                        &                      & 28.41       & 26.04  & 29.45  & 29.30  & 28.12  & {\ul 31.07}    & 30.27       & \textbf{31.92}      &                      & 0.8966        & 0.8986     & 0.8957               & 0.8954 & 0.8718 & {\ul 0.9240}  & 0.9233        & \textbf{0.9312}     \\
  $T_f \rightarrow T_2$                           &                      & 25.61       & 31.30  & 29.30  & 29.12  & 29.19  & {\ul 31.50}    & 30.78       & \textbf{32.83}      &                      & 0.8688        & 0.9107     & 0.8887               & 0.8851 & 0.8686 & {\ul 0.9285}  & 0.9234        & \textbf{0.9381}     \\ \hline
  AVERAGE                                         &                      & 30.08       & 31.91  & 31.48  & 30.64  & 30.53  & {\ul 33.72}    & 32.34       & \textbf{34.83}      &                      & 0.9074        & 0.9198     & 0.9047               & 0.9089 & 0.8824 & {\ul 0.9406}  & 0.9334        & \textbf{0.9466}     \\
                                                  &                      &             &        &        &        &        &                &             &                     & \multicolumn{1}{l}{} &               &            & \multicolumn{1}{l}{} &        &        &               &               &                     \\
  \multicolumn{1}{c}{\multirow{2}{*}{Modalities}} & \multicolumn{1}{c}{} & \multicolumn{8}{c}{rFID ↓}                                                                              & \multicolumn{1}{l}{} & \multicolumn{8}{c}{IS ↑}                                                                                                    \\
  \multicolumn{1}{c}{}                            & \multicolumn{1}{c}{} & MMSyn       & MMGAN  & MMT    & ModImp & M2DN   & Zhang          & CodeBrain   & \textbf{Ours} &                      & MMSyn         & MMGAN      & MMT                  & ModImp & M2DN   & Zhang         & CodeBrain     & \textbf{Ours} \\ \hline
  $T_1 \rightarrow T_{1c}$                        &                      & 52.25       & 118.12 & 112.16 & 112.75 & 134.75 & {\ul 27.49}    & 56.12       & \textbf{19.70}      &                      & {\ul 2.97}    & 2.47       & 2.52                 & 2.73   & 2.29   & 2.90          & 2.91          & \textbf{3.09}       \\
  $T_1 \rightarrow T_2$                           &                      & 38.85       & 113.48 & 98.18  & 155.30 & 93.75  & \textbf{14.38} & 23.37       & {\ul 18.92}         &                      & 1.73          & 1.95       & 1.93                 & 1.98   & 1.27   & \textbf{2.07} & 1.73          & {\ul 2.03}          \\
  $T_1 \rightarrow T_f$                           &                      & 39.16       & 129.69 & 132.42 & 141.37 & 94.71  & {\ul 31.41}    & 34.16       & \textbf{19.47}      &                      & 3.05          & 2.88       & 2.68                 & 2.30   & 2.39   & {\ul 3.15}    & \textbf{3.20} & 2.93                \\
  $T_{1c} \rightarrow T_1$                        &                      & 34.81       & 83.66  & 181.93 & 108.00 & 117.62 & 28.29          & {\ul 23.81} & \textbf{19.41}      &                      & {\ul 3.11}    & 2.74       & 2.75                 & 2.67   & 2.54   & 2.71          & 2.72          & \textbf{3.14}       \\
  $T_{1c} \rightarrow T_2$                        &                      & 43.23       & 98.51  & 115.51 & 160.10 & 94.79  & {\ul 29.70}    & 33.17       & \textbf{16.90}      &                      & 1.72          & 2.00       & \textbf{2.16}        & 1.91   & 1.25   & {\ul 2.11}    & 1.64          & 2.01                \\
  $T_{1c} \rightarrow T_f$                        &                      & 61.80       & 154.50 & 154.57 & 160.60 & 102.35 & 35.56          & {\ul 35.45} & \textbf{19.39}      &                      & 2.92          & 2.90       & 2.33                 & 2.39   & 2.26   & \textbf{3.03} & {\ul 3.02}    & 2.93                \\
  $T_2 \rightarrow T_1$                           &                      & 45.70       & 155.36 & 224.75 & 149.89 & 111.91 & \textbf{20.78} & {\ul 23.59} & 25.88               &                      & 2.36          & 2.96       & 2.97                 & 2.21   & 2.47   & {\ul 3.12}    & 2.98          & \textbf{3.13}       \\
  $T_2 \rightarrow T_{1c}$                        &                      & 76.88       & 146.39 & 171.22 & 167.69 & 143.21 & {\ul 36.53}    & 73.23       & \textbf{22.44}      &                      & 2.27          & 2.60       & 2.87                 & 2.44   & 2.09   & \textbf{3.11} & 2.91          & {\ul 3.08}          \\
  $T_2 \rightarrow T_f$                           &                      & 36.49       & 125.73 & 155.75 & 179.25 & 95.33  & {\ul 16.63}    & 27.83       & \textbf{16.34}      &                      & 2.75          & 2.89       & 2.73                 & 2.30   & 2.38   & \textbf{2.91} & 2.78          & 2.85                \\
  $T_f \rightarrow T_1$                           &                      & {\ul 43.33} & 131.76 & 179.63 & 183.45 & 121.98 & 55.86          & 58.35       & \textbf{22.50}      &                      & \textbf{3.33} & 2.76       & 3.13                 & 2.40   & 2.73   & 3.12          & 3.00          & {\ul 3.26}          \\
  $T_f \rightarrow T_{1c}$                        &                      & 66.66       & 152.82 & 124.95 & 177.32 & 133.72 & {\ul 34.11}    & 77.92       & \textbf{21.30}      &                      & \textbf{3.24} & 2.56       & 2.52                 & 2.40   & 2.36   & 3.10          & 2.87          & {\ul 3.21}          \\
  $T_f \rightarrow T_2$                           &                      & 43.79       & 129.58 & 105.97 & 178.85 & 94.69  & {\ul 29.17}    & 53.95       & \textbf{16.90}      &                      & 1.67          & 2.01       & \textbf{2.04}        & 1.79   & 1.29   & 1.72          & 1.43          & {\ul 2.01}          \\ \hline
  AVERAGE                                         &                      & 48.58       & 128.30 & 146.42 & 156.21 & 111.57 & {\ul 29.99}    & 43.41       & \textbf{19.93}      & \multicolumn{1}{l}{} & 2.59          & 2.56       & 2.55                 & 2.29   & 2.11   & {\ul 2.76}    & 2.60          & \textbf{2.81}
  \end{tabular}
  }
  \end{table}

\paragraph{N-to-N Synthesis.}
Table~\ref{tab:n2n_seg_summary} (left) summarises average performance over all $14$~input combinations (full per-combination results in Appendix~Table~\ref{app:tab_n2n}). \method{} leads on PSNR ($35.88$, $+0.75$\,dB over runner-up), SSIM ($0.9530$), and rFID ($19.68$, $-1.5\%$ vs.\ Zhang), with gains largest in single-source regimes. Simple latent averaging suffices for multi-source fusion because all modalities share the same hypersphere.

\begin{table}[t]
    \centering
    \caption{N-to-N synthesis and segmentation averages on BraTS (14 input combinations). Full oracle = all ground-truth modalities. Per-combination results in Appendix Tables~\ref{app:tab_n2n},~\ref{app:tab_seg}.}
    \label{tab:n2n_seg_summary}
    \resizebox{0.78\textwidth}{!}{
    \begin{tabular}{lcccccccccc}
    \toprule
    & \multicolumn{4}{c}{\textbf{Synthesis (N-to-N avg)}} & & \multicolumn{4}{c}{\textbf{Segmentation (avg)}} \\
    \cmidrule{2-5} \cmidrule{7-10}
    Method & PSNR~$\uparrow$ & SSIM~$\uparrow$ & rFID~$\downarrow$ & IS~$\uparrow$ & & Dice~$\uparrow$ & HD95~$\downarrow$ & IoU~$\uparrow$ & Sens~$\uparrow$ \\
    \midrule
    Zero-fill & --- & --- & --- & --- & & 0.3185 & 169.40 & 0.2693 & 0.3260 \\
    MMSyn & 31.81 & 0.9213 & 37.12 & 2.64 & & 0.5546 & 57.73 & 0.4641 & 0.5553 \\
    MMGAN & 33.81 & 0.9339 & 93.12 & 2.51 & & 0.5657 & 55.39 & 0.4877 & 0.5524 \\
    MMT & 32.40 & 0.9125 & 135.77 & 2.48 & & 0.5450 & {\ul 46.27} & 0.4633 & 0.5374 \\
    ModImp & 32.23 & 0.9226 & 123.87 & 2.42 & & 0.5627 & 59.44 & 0.4832 & 0.5574 \\
    M2DN & 31.68 & 0.8936 & 104.17 & 2.18 & & 0.5393 & 53.32 & 0.4507 & 0.5028 \\
    Zhang & {\ul 35.13} & {\ul 0.9504} & {\ul 19.98} & {\ul 2.76} & & 0.5080 & 53.06 & 0.4216 & 0.4872 \\
    CodeBrain & 33.88 & 0.9446 & 32.45 & 2.63 & & {\ul 0.5930} & 56.37 & {\ul 0.5174} & {\ul 0.5759} \\
    \textbf{Ours} & \textbf{35.88} & \textbf{0.9530} & \textbf{19.68} & \textbf{2.80} & & \textbf{0.7473} & \textbf{19.02} & \textbf{0.6511} & \textbf{0.7418} \\
    \midrule
    Full oracle & --- & --- & --- & --- & & 0.8554 & 11.41 & 0.7829 & 0.8583 \\
    \bottomrule
    \end{tabular}
    }
\end{table}

\vspace{-2pt}
\subsection{Cross-Domain Translation: CT \boldmath$\leftrightarrow$ MRI}
\label{sec:exp_ctmri}
\vspace{-2pt}

CT--MRI translation is harder than within-MRI synthesis due to fundamentally different imaging physics~\citep{wang2024softmamba}. Table~\ref{tab:ctmri_metrics} shows that \method{} achieves the best PSNR/SSIM/rFID on $\mathrm{MRI}{\to}\mathrm{CT}$ ($27.82$\,dB, rFID~$13.99$). On $\mathrm{CT}{\to}\mathrm{MRI}$, Zhang slightly leads PSNR/SSIM, but \method{} halves rFID ($30.91{\to}13.65$), reflecting the geometric advantage of decoding on-manifold latents.

\paragraph{Downstream Segmentation.}\label{sec:exp_seg}
Table~\ref{tab:n2n_seg_summary} (right) evaluates downstream segmentation: a 3D~U-Net trained on full-modality BraTS and tested on imputed inputs (full per-combination results in Appendix~Table~\ref{app:tab_seg}). \method{} reaches avg Dice $0.7473$ ($+15.4$ over runner-up CodeBrain, within $0.108$ of the full-modality oracle $0.8554$) and reduces HD95 from $46.27$ to $19.02$ ($-59\%$). In single-source cases, baselines collapse below Dice $0.10$--$0.20$, while \method{} stays above $0.45$.

\begin{table}[t]
\begin{minipage}[t]{0.58\textwidth}
\centering
\captionof{table}{CT $\leftrightarrow$ MRI translation on SynthRAD2023.}
\label{tab:ctmri_metrics}
\resizebox{\textwidth}{!}{
    \begin{tabular}{ccccccccccc}
  \multirow{2}{*}{Methods} &  & \multicolumn{4}{c}{$CT \rightarrow MRI$}                          &           & \multicolumn{4}{c}{$MRI \rightarrow CT$}                          \\
                           &  & PSNR ↑           & SSIM ↑            & rFID ↓           & IS ↑            &           & PSNR ↑           & SSIM ↑            & rFID ↓           & IS ↑            \\ \hline
  MMSyn                    &  & 21.42          & 0.7666          & 107.89         & 2.98          &           & 22.56          & 0.8112          & 98.76          & 2.73          \\
  MMGAN                    &  & 22.58          & 0.7543          & 182.83         & 2.95          &           & 25.69          & 0.7983          & 131.40         & 2.52          \\
  ModImp                 &  & 16.40          & 0.6535          & 217.59         & 2.72          &           & 25.30          & 0.7553          & 140.33         & 2.46          \\
  MMT                      &  & 23.68          & 0.7488          & 213.48         & 2.65          &           & 25.40          & 0.8105          & 145.04         & 2.32          \\
  CodeBrain                &  & 22.94          & 0.8083          & 54.82          & \textbf{3.08} & \textbf{} & 25.26          & 0.8569          & 61.31          & 2.76          \\
  Zhang                    &  & \textbf{25.51} & \textbf{0.8635} & {\ul 30.91}    & {\ul 3.01}    & {\ul }    & {\ul 27.52}    & {\ul 0.8944}    & {\ul 38.61}    & \textbf{2.92} \\
  M2DN                     &  & 21.44          & 0.7497          & 117.02         & 2.42          &           & 23.79          & 0.8017          & 99.15          & {\ul 2.88}    \\ \hline
  \textbf{Ours}      &  & {\ul 24.70}    & {\ul 0.8452}    & \textbf{13.65} & 2.71          &           & \textbf{27.82} & \textbf{0.9084} & \textbf{13.99} & 2.81
  \end{tabular}
  }
\end{minipage}\hfill
\begin{minipage}[t]{0.40\textwidth}
\centering
\captionof{table}{Encoder comparison (Stage-1 reconstruction on $T_{1c}$).}
\label{tab:rec_encoders}
\resizebox{\textwidth}{!}{
  \begin{tabular}{ccccccc}
  \multirow{2}{*}{Encoders} &  & \multicolumn{5}{c}{$T_{1c}$}                                                            \\ \cline{3-7}
                            &  & PSNR ↑           & SSIM ↑            & rFID ↓           & IS ↑            & iFID ↓           \\ \hline
  MedSAM                    &  & 29.17          & 0.8642          & 140.82         & 1.96          & 48.58          \\
  MedMAE                    &  & \textbf{31.23} & \textbf{0.8907} & 43.17          & 2.84          & 144.45         \\
  DINOv2                    &  & 30.57          & 0.8824          & \textbf{35.29} & \textbf{2.91} & \textbf{40.64}
  \end{tabular}
  }
\end{minipage}
\end{table}

\vspace{-2pt}
\subsection{Ablation Studies}
\label{sec:exp_ablation}
\vspace{-2pt}

\paragraph{Choice of Encoder.}
Table~\ref{tab:rec_encoders} compares encoders at Stage-1 reconstruction on $T_{1c}$. MedMAE wins PSNR ($31.23$) but suffers $3.6\times$ higher iFID ($144.45$ vs.\ $40.64$); DINOv2 trades $0.66$\,dB for the best rFID/IS/iFID, confirming that geometric regularity matters more than pixel fidelity for transport-based imputation.

\paragraph{Stage-wise RAE Ablation.}
We progressively enable each Stage-1 step and evaluate on-manifold ($z{=}\mathcal{E}_{\text{DINO}}(x_t)$) and chord ($z_{1/2}{=}\tfrac{1}{2}(z_s{+}z_t)$) reconstruction.
\vspace{-2pt}
\begin{center}
\captionof{table}{Stage-wise RAE ablation ($T_1$/$T_{1c}$).}\label{tab:rae_ablation}
\resizebox{0.72\textwidth}{!}{
\begin{tabular}{lcccccc}
    \hline
    \multirow{2}{*}{Cascade stage} & \multicolumn{3}{c}{On-manifold} & \multicolumn{3}{c}{Chord ($\tau{=}0.5$)} \\
                                   & PSNR~$\uparrow$ & SSIM~$\uparrow$ & rFID~$\downarrow$ & PSNR~$\uparrow$ & SSIM~$\uparrow$ & rFID~$\downarrow$ \\ \hline
    Step~1: single-modality warm-up        & 33.14 & 0.8732 & 28.29           & 32.04 & 0.8731 & 41.66           \\
    \quad+ Step~2: manifold expansion      & 33.83 & 0.8795 & 30.92           & 33.29 & 0.8784 & 35.06           \\
    \quad+ Step~3: shared-encoder align    & \textbf{44.06} & \textbf{0.9545} & \textbf{16.55} & \textbf{41.40} & \textbf{0.9453} & \textbf{16.28} \\
    \hline
\end{tabular}
}
\end{center}
\vspace{-6pt}
Table~\ref{tab:rae_ablation} shows that Step~2 (manifold expansion) provides a modest gain on its own ($+0.69$\,dB on-manifold, $+1.25$\,dB chord) by exposing the decoder to off-manifold inputs, but the chord rFID actually worsens slightly ($28.29{\to}30.92$), indicating that the decoder generalizes to nearby off-manifold points without yet aligning modality-specific representations. Step~3 (shared-encoder alignment) is decisive: on-manifold PSNR jumps to $44.06$ and the chord-vs-on-manifold rFID gap collapses from $13.4$ to $0.3$, confirming that once all modality decoders share a geometrically consistent latent neighborhood, the flow trajectory becomes nearly indistinguishable from the data manifold.


\vspace{-2pt}
\section{Conclusion}
\label{sec:conclusion}
\vspace{-2pt}

We presented \method{}, which reframes missing modality imputation as short-range transport on a shared hypersphere~$\mathcal{S}^{d-1}$, replacing the noise source with the observed-modality latent and expanding the decoder domain to cover the off-manifold region the flow traverses.
These two geometry-aware modifications reduce the worst-case off-manifold deviation from~$\Theta(\sqrt{d}\,)$ to a dimension-independent bound~$1{-}\cos(\alpha/2)$, and yield state-of-the-art fidelity on BraTS and CT--MRI translation while closing roughly $80\%$ of the segmentation gap to the full-modality oracle.

\section*{Limitation}
A current limitation is the reliance on DINOv2 providing cross-modal compactness; modalities without shared visual semantics (e.g., text and imaging) would require a domain-adapted encoder or a cross-modal pre-training stage.
More broadly, the principle of aligning the latent geometry with the task structure should extend to other conditional generation settings where paired source data is available, such as image-guided speech synthesis or satellite multi-spectral fusion.

\begin{ack}
This work was partially supported by the RGC Collaborative Research Fund (No.~C5055-24G), the Seed Fund of the Research Institute for Smart Ageing (No.~P0050946), the Tsinghua-PolyU Joint Research Initiative Fund (No.~P0056509), and PolyU UGC funding (No.~P0053716).
The authors declare no competing interests.
\end{ack}


\bibliographystyle{plainnat}
\bibliography{ref}

@article{brats2021,
  title={The rsna-asnr-miccai brats 2021 benchmark on brain tumor segmentation and radiogenomic classification},
  author={Baid, Ujjwal and Ghodasara, Satyam and Mohan, Suyash and Bilello, Michel and Calabrese, Evan and Colak, Errol and Farahani, Keyvan and Kalpathy-Cramer, Jayashree and Kitamura, Felipe C and Pati, Sarthak and others},
  journal={arXiv preprint arXiv:2107.02314},
  year={2021}
}

@article{chartsias2018multimodal,
  title={Multimodal MR synthesis via modality-invariant latent representation},
  author={Chartsias, Agisilaos and Joyce, Thomas and Giuffrida, Mario Valerio and Tsaftaris, Sotirios A},
  journal={IEEE transactions on medical imaging},
  volume={37},
  number={3},
  pages={803--814},
  year={2017},
  publisher={IEEE}
}

@article{chen2023riemannian,
  title={Flow matching on general geometries},
  author={Chen, Ricky TQ and Lipman, Yaron},
  journal={arXiv preprint arXiv:2302.03660},
  year={2023}
}

@inproceedings{peebles2023dit,
  title={Scalable diffusion models with transformers},
  author={Peebles, William and Xie, Saining},
  booktitle={Proceedings of the IEEE/CVF international conference on computer vision},
  pages={4195--4205},
  year={2023}
}

@article{dar2019image,
  title={Image synthesis in multi-contrast MRI with conditional generative adversarial networks},
  author={Dar, Salman UH and Yurt, Mahmut and Karacan, Levent and Erdem, Aykut and Erdem, Erkut and Cukur, Tolga},
  journal={IEEE transactions on medical imaging},
  volume={38},
  number={10},
  pages={2375--2388},
  year={2019},
  publisher={IEEE}
}

@article{dinosae2026,
  title={DINO-SAE: DINO Spherical Autoencoder for High-Fidelity Image Reconstruction and Generation},
  author={Chang, Hun and Cha, Byunghee and Ye, Jong Chul},
  journal={arXiv preprint arXiv:2601.22904},
  year={2026}
}

@inproceedings{he2022masked,
  title={Masked autoencoders are scalable vision learners},
  author={He, Kaiming and Chen, Xinlei and Xie, Saining and Li, Yanghao and Doll{\'a}r, Piotr and Girshick, Ross},
  booktitle={Proceedings of the IEEE/CVF conference on computer vision and pattern recognition},
  pages={16000--16009},
  year={2022}
}

@article{kingma2014auto,
  title={Auto-encoding variational bayes},
  author={Kingma, Diederik P and Welling, Max},
  journal={arXiv preprint arXiv:1312.6114},
  year={2013}
}

@article{kumar2026manifold,
  title={Learning on the Manifold: Unlocking Standard Diffusion Transformers with Representation Encoders},
  author={Kumar, Amandeep and Patel, Vishal M},
  journal={arXiv preprint arXiv:2602.10099},
  year={2026}
}

@article{lipman2023flow,
  title={Flow matching for generative modeling},
  author={Lipman, Yaron and Chen, Ricky TQ and Ben-Hamu, Heli and Nickel, Maximilian and Le, Matt},
  journal={arXiv preprint arXiv:2210.02747},
  year={2022}
}

@article{lvrae2026,
  title={Improving Reconstruction of Representation Autoencoder},
  author={Liu, Siyu and Qin, Chujie and Yin, Hubery and Yan, Qixin and Duan, Zheng-Peng and Li, Chen and Lyu, Jing and Guo, Chun-Le and Li, Chongyi},
  journal={arXiv preprint arXiv:2602.08620},
  year={2026}
}

@inproceedings{misaldm2025,
  title={Structure-Aware MRI Translation: Multi-modal Latent Diffusion Model with Arbitrary Missing Modalities},
  author={Zhang, Xinzhe and Liang, Junjie and Cao, Peng and Yang, Jinzhu and Zaiane, Osmar R},
  booktitle={International Conference on Medical Image Computing and Computer-Assisted Intervention},
  pages={508--518},
  year={2025},
  organization={Springer}
}

@article{liu2023mmt,
  title={One model to synthesize them all: Multi-contrast multi-scale transformer for missing data imputation},
  author={Liu, Jiang and Pasumarthi, Srivathsa and Duffy, Ben and Gong, Enhao and Datta, Keshav and Zaharchuk, Greg},
  journal={IEEE transactions on medical imaging},
  volume={42},
  number={9},
  pages={2577--2591},
  year={2023},
  publisher={IEEE}
}

@article{meng2024m2dn,
  title={Multi-modal modality-masked diffusion network for brain mri synthesis with random modality missing},
  author={Meng, Xiangxi and Sun, Kaicong and Xu, Jun and He, Xuming and Shen, Dinggang},
  journal={IEEE Transactions on Medical Imaging},
  volume={43},
  number={7},
  pages={2587--2598},
  year={2024},
  publisher={IEEE}
}

@article{wu2025codebrain,
  title={CodeBrain: Imputing Any Brain MRI via Modality-and Instance-Specific Codes},
  author={Wu, Yicheng and Song, Tao and Wu, Zhonghua and Ye, Jin and Ge, Zongyuan and Chen, Zhaolin and Cai, Jianfei},
  journal={arXiv preprint arXiv:2501.18328},
  year={2025}
}

@inproceedings{qi2025modimp,
  title={A Unified Missing Modality Imputation Model with Inter-modality Contrastive and Consistent Learning},
  author={Qi, Liangce and Liu, Yusi and Li, Yuqin and Shi, Weili and Feng, Guanyuan and Jiang, Zhengang},
  booktitle={International Conference on Medical Image Computing and Computer-Assisted Intervention},
  pages={44--53},
  year={2025},
  organization={Springer}
}

@article{zhang2024unified,
  title={Unified multi-modal image synthesis for missing modality imputation},
  author={Zhang, Yue and Peng, Chengtao and Wang, Qiuli and Song, Dan and Li, Kaiyan and Zhou, S Kevin},
  journal={IEEE Transactions on Medical Imaging},
  volume={44},
  number={1},
  pages={4--18},
  year={2024},
  publisher={IEEE}
}

@article{synthrad2023,
  title={SynthRAD2023 Grand Challenge dataset: Generating synthetic CT for radiotherapy},
  author={Thummerer, Adrian and Van der Bijl, Erik and Galapon Jr, Arthur and Verhoeff, Joost JC and Langendijk, Johannes A and Both, Stefan and van den Berg, Cornelis (Nico) AT and Maspero, Matteo},
  journal={Medical physics},
  volume={50},
  number={7},
  pages={4664--4674},
  year={2023},
  publisher={Wiley Online Library}
}

@article{wang2004ssim,
  title={Image quality assessment: from error visibility to structural similarity},
  author={Wang, Zhou and Bovik, Alan C and Sheikh, Hamid R and Simoncelli, Eero P},
  journal={IEEE transactions on image processing},
  volume={13},
  number={4},
  pages={600--612},
  year={2004},
  publisher={IEEE}
}

@article{heusel2017fid,
  title={Gans trained by a two time-scale update rule converge to a local nash equilibrium},
  author={Heusel, Martin and Ramsauer, Hubert and Unterthiner, Thomas and Nessler, Bernhard and Hochreiter, Sepp},
  journal={Advances in neural information processing systems},
  volume={30},
  year={2017}
}

@article{salimans2016is,
  title={Improved techniques for training gans},
  author={Salimans, Tim and Goodfellow, Ian and Zaremba, Wojciech and Cheung, Vicki and Radford, Alec and Chen, Xi},
  journal={Advances in neural information processing systems},
  volume={29},
  year={2016}
}

@article{xu2026ifid,
  title={Making Reconstruction FID Predictive of Diffusion Generation FID},
  author={Xu, Tongda and He, Mingwei and Abu-Hussein, Shady and Hernandez-Lobato, Jose Miguel and Zhang, Haotian and Zhao, Kai and Zhou, Chao and Zhang, Ya-Qin and Wang, Yan},
  journal={arXiv preprint arXiv:2603.05630},
  year={2026}
}

@article{oquab2024dinov2,
  title={Dinov2: Learning robust visual features without supervision},
  author={Oquab, Maxime and Darcet, Timoth{\'e}e and Moutakanni, Th{\'e}o and Vo, Huy and Szafraniec, Marc and Khalidov, Vasil and Fernandez, Pierre and Haziza, Daniel and Massa, Francisco and El-Nouby, Alaaeldin and others},
  journal={arXiv preprint arXiv:2304.07193},
  year={2023}
}

@inproceedings{rombach2022high,
  title={High-resolution image synthesis with latent diffusion models},
  author={Rombach, Robin and Blattmann, Andreas and Lorenz, Dominik and Esser, Patrick and Ommer, Bj{\"o}rn},
  booktitle={Proceedings of the IEEE/CVF conference on computer vision and pattern recognition},
  pages={10684--10695},
  year={2022}
}

@article{sharma2019missing,
  title={Missing MRI pulse sequence synthesis using multi-modal generative adversarial network},
  author={Sharma, Anmol and Hamarneh, Ghassan},
  journal={IEEE transactions on medical imaging},
  volume={39},
  number={4},
  pages={1170--1183},
  year={2019},
  publisher={IEEE}
}

@article{zheng2025rae,
  title={Diffusion transformers with representation autoencoders},
  author={Zheng, Boyang and Ma, Nanye and Tong, Shengbang and Xie, Saining},
  journal={arXiv preprint arXiv:2510.11690},
  year={2025}
}

@article{wang2021medical,
  title={A review on medical imaging synthesis using deep learning and its clinical applications},
  author={Wang, Tonghe and Lei, Yang and Fu, Yabo and Wynne, Jacob F and Curran, Walter J and Liu, Tian and Yang, Xiaofeng},
  journal={Journal of applied clinical medical physics},
  volume={22},
  number={1},
  pages={11--36},
  year={2021},
  publisher={Wiley Online Library}
}

@article{goodfellow2014gan,
  title={Generative adversarial nets},
  author={Goodfellow, Ian J and Pouget-Abadie, Jean and Mirza, Mehdi and Xu, Bing and Warde-Farley, David and Ozair, Sherjil and Courville, Aaron and Bengio, Yoshua},
  journal={Advances in neural information processing systems},
  volume={27},
  year={2014}
}

@inproceedings{isola2017pix2pix,
  title={Image-to-image translation with conditional adversarial networks},
  author={Isola, Phillip and Zhu, Jun-Yan and Zhou, Tinghui and Efros, Alexei A},
  booktitle={Proceedings of the IEEE conference on computer vision and pattern recognition},
  pages={1125--1134},
  year={2017}
}

@inproceedings{zhu2017cyclegan,
  title={Unpaired image-to-image translation using cycle-consistent adversarial networks},
  author={Zhu, Jun-Yan and Park, Taesung and Isola, Phillip and Efros, Alexei A},
  booktitle={Proceedings of the IEEE international conference on computer vision},
  pages={2223--2232},
  year={2017}
}

@article{vaswani2017attention,
  title={Attention is all you need},
  author={Vaswani, Ashish and Shazeer, Noam and Parmar, Niki and Uszkoreit, Jakob and Jones, Llion and Gomez, Aidan N and Kaiser, {\L}ukasz and Polosukhin, Illia},
  journal={Advances in neural information processing systems},
  volume={30},
  year={2017}
}

@article{dosovitskiy2021vit,
  title={An image is worth 16x16 words: Transformers for image recognition at scale},
  author={Dosovitskiy, Alexey and Beyer, Lucas and Kolesnikov, Alexander and Weissenborn, Dirk and Zhai, Xiaohua and Unterthiner, Thomas and Dehghani, Mostafa and Minderer, Matthias and Heigold, Georg and Gelly, Sylvain and others},
  journal={arXiv preprint arXiv:2010.11929},
  year={2020}
}

@article{ho2020ddpm,
  title={Denoising diffusion probabilistic models},
  author={Ho, Jonathan and Jain, Ajay and Abbeel, Pieter},
  journal={Advances in neural information processing systems},
  volume={33},
  pages={6840--6851},
  year={2020}
}

@article{song2021sde,
  title={Score-based generative modeling through stochastic differential equations},
  author={Song, Yang and Sohl-Dickstein, Jascha and Kingma, Diederik P and Kumar, Abhishek and Ermon, Stefano and Poole, Ben},
  journal={arXiv preprint arXiv:2011.13456},
  year={2020}
}

@article{liu2022rectified,
  title={Flow straight and fast: Learning to generate and transfer data with rectified flow},
  author={Liu, Xingchao and Gong, Chengyue and Liu, Qiang},
  journal={arXiv preprint arXiv:2209.03003},
  year={2022}
}

@inproceedings{esser2024sd3,
  title={Scaling rectified flow transformers for high-resolution image synthesis},
  author={Esser, Patrick and Kulal, Sumith and Blattmann, Andreas and Entezari, Rahim and M{\"u}ller, Jonas and Saini, Harry and Levi, Yam and Lorenz, Dominik and Sauer, Axel and Boesel, Frederic and others},
  booktitle={Forty-first international conference on machine learning},
  year={2024}
}

@inproceedings{chen2020simclr,
  title={A simple framework for contrastive learning of visual representations},
  author={Chen, Ting and Kornblith, Simon and Norouzi, Mohammad and Hinton, Geoffrey},
  booktitle={International conference on machine learning},
  pages={1597--1607},
  year={2020},
  organization={PmLR}
}

@inproceedings{he2020moco,
  title={Momentum contrast for unsupervised visual representation learning},
  author={He, Kaiming and Fan, Haoqi and Wu, Yuxin and Xie, Saining and Girshick, Ross},
  booktitle={Proceedings of the IEEE/CVF conference on computer vision and pattern recognition},
  pages={9729--9738},
  year={2020}
}

@inproceedings{caron2021dino,
  title={Emerging properties in self-supervised vision transformers},
  author={Caron, Mathilde and Touvron, Hugo and Misra, Ishan and J{\'e}gou, Herv{\'e} and Mairal, Julien and Bojanowski, Piotr and Joulin, Armand},
  booktitle={Proceedings of the IEEE/CVF international conference on computer vision},
  pages={9650--9660},
  year={2021}
}

@inproceedings{hore2010psnr,
  title={Image quality metrics: PSNR vs. SSIM},
  author={Hore, Alain and Ziou, Djemel},
  booktitle={2010 20th international conference on pattern recognition},
  pages={2366--2369},
  year={2010},
  organization={IEEE}
}

@article{wang2024softmamba,
  title={Soft masked mamba diffusion model for ct to mri conversion},
  author={Wang, Zhenbin and Zhang, Lei and Wang, Lituan and Zhang, Zhenwei},
  journal={arXiv preprint arXiv:2406.15910},
  year={2024}
}

\clearpage
\appendix
\section{Proof of Proposition~\ref{prop:bound}}
\label{app:proof_bound}

\begin{proof}
We proceed in three parts: (i)~we derive $\|z_\tau\|^2$ as a function of~$\tau$, (ii)~we minimize it over~$[0,1]$ to identify the point of maximum off-manifold deviation, and (iii)~we compare with the Gaussian-source setting.

\paragraph{Part~(i): Norm of the interpolant.}
Let $z_s, z_t \in \mathcal{S}^{d-1}$ with inner product $\langle z_s, z_t \rangle = \cos\alpha$ for angular separation $\alpha \in [0, \pi]$. The linear interpolant is $z_\tau = \tau\, z_t + (1{-}\tau)\, z_s$ for $\tau \in [0,1]$. Expanding the squared norm:
\begin{align}
    \|z_\tau\|^2
    &= \|\tau\, z_t + (1{-}\tau)\, z_s\|^2 \notag \\
    &= \tau^2 \|z_t\|^2 + (1{-}\tau)^2 \|z_s\|^2 + 2\tau(1{-}\tau) \langle z_s, z_t \rangle \notag \\
    &= \tau^2 + (1{-}\tau)^2 + 2\tau(1{-}\tau)\cos\alpha
    \label{eq:app_norm_sq}
\end{align}
where the last equality uses $\|z_s\| = \|z_t\| = 1$.
Expanding and collecting terms in~$\tau$:
\begin{equation}
    \|z_\tau\|^2 = 2(1-\cos\alpha)\,\tau^2 - 2(1-\cos\alpha)\,\tau + 1
    \label{eq:app_quadratic}
\end{equation}
This is a quadratic in~$\tau$ with leading coefficient $a \coloneqq 2(1 - \cos\alpha) \geq 0$.

\paragraph{Part~(ii): Minimizing $\|z_\tau\|^2$ over $[0,1]$.}
\emph{Case $\alpha = 0$} ($z_s = z_t$). The coefficient $a = 0$, so $\|z_\tau\|^2 = 1$ for all~$\tau$, and $\Delta = 0$ trivially.

\emph{Case $\alpha \in (0, \pi]$}. Now $a > 0$, so~\eqref{eq:app_quadratic} is a strictly convex quadratic in~$\tau$.
Its unique global minimizer is
\begin{equation}
    \tau^{\star} = \frac{2(1-\cos\alpha)}{2 \cdot 2(1-\cos\alpha)} = \frac{1}{2}
\end{equation}
Since $\tau^{\star} = \tfrac{1}{2} \in [0,1]$, the unconstrained minimizer lies within the domain and is therefore the constrained minimizer as well.
Substituting $\tau = \tfrac{1}{2}$ into~\eqref{eq:app_quadratic}:
\begin{equation}
    \|z_{1/2}\|^2
    = 2(1-\cos\alpha) \cdot \tfrac{1}{4} - 2(1-\cos\alpha) \cdot \tfrac{1}{2} + 1
    = 1 - \tfrac{1}{2}(1-\cos\alpha)
    = \frac{1+\cos\alpha}{2}
    \label{eq:app_min_norm}
\end{equation}

We now apply the half-angle identity $\cos^2(\alpha/2) = (1+\cos\alpha)/2$.
Since $\alpha \in [0,\pi]$ we have $\alpha/2 \in [0,\pi/2]$, so $\cos(\alpha/2) \geq 0$ and hence
\begin{equation}
    \|z_{1/2}\| = \sqrt{\frac{1+\cos\alpha}{2}} = \sqrt{\cos^2\!\frac{\alpha}{2}} = \cos\frac{\alpha}{2}
\end{equation}

Because $\|z_\tau\|^2$ is minimized at $\tau = \tfrac{1}{2}$, so is $\|z_\tau\|$ (the square root is monotone), and the deviation $1 - \|z_\tau\|$ is maximized there.
To confirm that the deviation is non-negative, rewrite~\eqref{eq:app_norm_sq} as $\|z_\tau\|^2 = 1 - 2\tau(1{-}\tau)(1{-}\cos\alpha)$.
Since $\tau(1{-}\tau) \geq 0$ for $\tau \in [0,1]$ and $1 - \cos\alpha \geq 0$, the subtracted term is non-negative, so $\|z_\tau\|^2 \leq 1$ and hence $\|z_\tau\| \leq 1$ for all $\tau \in [0,1]$.
Therefore:
\begin{equation}
    \Delta
    \;\coloneqq\; \max_{\tau \in [0,1]} \bigl(1 - \|z_\tau\|\bigr)
    = 1 - \|z_{1/2}\|
    = 1 - \cos\frac{\alpha}{2}
\end{equation}

\paragraph{Part~(iii): Gaussian-source comparison.}
In standard flow matching the source is $\varepsilon \sim \mathcal{N}(0, I_d)$ with interpolant $z_\tau = \tau\,z + (1{-}\tau)\,\varepsilon$, where $z \in \mathcal{S}^{d-1}$.
We show $\mathbb{E}\!\bigl[\max_{\tau \in [0,1]} \bigl|\,\|z_\tau\| - 1\,\bigr|\bigr] = \Theta(\sqrt{d}\,)$.
Write $\mu \coloneqq \mathbb{E}[\|\varepsilon\|] = \sqrt{2}\,\Gamma\!\bigl(\tfrac{d+1}{2}\bigr)\big/\Gamma\!\bigl(\tfrac{d}{2}\bigr) = \sqrt{d} - \tfrac{1}{4\sqrt{d}} + O(d^{-3/2})$.

\emph{Lower bound.}
At $\tau = 0$, $z_0 = \varepsilon$, so $\max_{\tau} \bigl|\,\|z_\tau\| - 1\,\bigr| \geq \|\varepsilon\| - 1$ whenever $\|\varepsilon\| \geq 1$.
Since $\|\varepsilon\|^2 \sim \chi^2_d$, the event $\|\varepsilon\| < 1$ has probability at most $e^{-\Omega(d)}$ for $d \geq 2$.
Therefore
\begin{equation}
    \mathbb{E}\!\Bigl[\max_{\tau} \bigl|\,\|z_\tau\| - 1\,\bigr|\Bigr]
    \;\geq\; \mu - 1 - O(e^{-\Omega(d)})
    = \Omega(\sqrt{d}\,)
\end{equation}

\emph{Upper bound.}
By the triangle inequality, $\|z_\tau\| \leq \tau\|z\| + (1{-}\tau)\|\varepsilon\| = \tau + (1{-}\tau)\|\varepsilon\|$ for all~$\tau$.
Whenever $\|\varepsilon\| \geq 2$, two cases exhaust the possibilities for every~$\tau \in [0,1]$:
if $\|z_\tau\| \geq 1$, then $\bigl|\,\|z_\tau\| - 1\bigr| = \|z_\tau\| - 1 \leq (1{-}\tau)(\|\varepsilon\| - 1) \leq \|\varepsilon\| - 1$;
if $\|z_\tau\| < 1$, then $\bigl|\,\|z_\tau\| - 1\bigr| = 1 - \|z_\tau\| \leq 1 \leq \|\varepsilon\| - 1$.
Hence $\max_\tau \bigl|\,\|z_\tau\| - 1\bigr| \leq \|\varepsilon\| - 1$ whenever $\|\varepsilon\| \geq 2$.
Since $\Pr[\|\varepsilon\| < 2] = e^{-\Omega(d)}$ for large~$d$:
\begin{equation}
    \mathbb{E}\!\Bigl[\max_{\tau} \bigl|\,\|z_\tau\| - 1\,\bigr|\Bigr]
    \;\leq\; \mu - 1 + O(e^{-\Omega(d)})
    = O(\sqrt{d}\,)
\end{equation}

Combining: $\mathbb{E}\!\bigl[\max_{\tau} \bigl|\,\|z_\tau\| - 1\,\bigr|\bigr] = \Theta(\sqrt{d}\,)$, which grows with dimension.
By contrast, the data-as-source deviation $\Delta = 1 - \cos(\alpha/2)$ depends only on the angular separation~$\alpha$ and is independent of~$d$.
(The two notions of deviation coincide: as shown in Part~(ii), $\|z_\tau\| \leq 1$ when both endpoints lie on $\mathcal{S}^{d-1}$, so $1 - \|z_\tau\| = \bigl|\,\|z_\tau\| - 1\,\bigr|$ throughout.)
\end{proof}

\section{Evaluation Metrics}
\label{app:metrics}

We employ two families of metrics: \emph{generation quality} metrics that compare each synthesized image to its ground-truth counterpart, and \emph{downstream segmentation} metrics that measure how well a pre-trained segmentation model performs when missing modalities are filled in by each imputation method.

\subsection{Generation Quality Metrics}

\paragraph{Peak Signal-to-Noise Ratio (PSNR)~\cite{hore2010psnr}.}
PSNR measures per-pixel reconstruction fidelity between a synthesized image~$\hat{x}$ and its ground truth~$x$, defined as
\begin{equation}
    \text{PSNR}(\hat{x}, x) = 10 \cdot \log_{10} \frac{\text{MAX}^2}{\text{MSE}(\hat{x}, x)}
\end{equation}
where $\text{MAX}$ is the maximum possible pixel value and $\text{MSE}(\hat{x}, x) = \frac{1}{N}\sum_{i=1}^{N}(\hat{x}_i - x_i)^2$ is the mean squared error over all $N$~pixels. Higher is better. We report PSNR in decibels (dB), averaged over all test slices.

\paragraph{Structural Similarity Index (SSIM)~\cite{wang2004ssim}.}
SSIM captures perceptual image quality by comparing local luminance, contrast, and structure between $\hat{x}$ and $x$:
\begin{equation}
    \text{SSIM}(\hat{x}, x) = \frac{(2\mu_{\hat{x}}\mu_x + C_1)(2\sigma_{\hat{x}x} + C_2)}{(\mu_{\hat{x}}^2 + \mu_x^2 + C_1)(\sigma_{\hat{x}}^2 + \sigma_x^2 + C_2)}
\end{equation}
where $\mu$ and $\sigma^2$ denote local means and variances computed over a sliding window, $\sigma_{\hat{x}x}$ is the local cross-covariance, and $C_1, C_2$ are small stabilizing constants. SSIM ranges in $[-1, 1]$; higher is better. We use the default $11 \times 11$ Gaussian window and report the mean over all test slices.

\paragraph{Reconstruction Fr\'echet Inception Distance (rFID)~\cite{heusel2017fid}.}
rFID measures distributional similarity between the set of synthesized images and the set of ground-truth images in Inception-v3 feature space. It is computed as the Fr\'echet distance between two multivariate Gaussians fitted to the Inception features:
\begin{equation}
    \text{rFID} = \|\mu_{\hat{x}} - \mu_x\|^2 + \text{Tr}\!\left(\Sigma_{\hat{x}} + \Sigma_x - 2(\Sigma_{\hat{x}}\Sigma_x)^{1/2}\right)
\end{equation}
where $(\mu_{\hat{x}}, \Sigma_{\hat{x}})$ and $(\mu_x, \Sigma_x)$ are the mean and covariance of the Inception features for the synthesized and real sets, respectively. Lower is better. Following standard practice, we compute rFID in a \emph{paired} fashion: the synthesized set and ground-truth set contain the same subjects, so rFID captures both sample quality and alignment fidelity. We compute rFID per imputation direction and report the average.

\paragraph{Inception Score (IS)~\cite{salimans2016is}.}
IS evaluates the quality and diversity of generated images using the Inception-v3 classifier:
\begin{equation}
    \text{IS} = \exp\!\left(\mathbb{E}_{\hat{x}}\left[D_{\text{KL}}\!\left(p(y|\hat{x}) \,\|\, p(y)\right)\right]\right)
\end{equation}
where $p(y|\hat{x})$ is the conditional class distribution predicted by Inception for a synthesized image, and $p(y) = \mathbb{E}_{\hat{x}}[p(y|\hat{x})]$ is the marginal class distribution. A high IS indicates that individual images are confidently classified (sharp $p(y|\hat{x})$) while the overall set is diverse (broad $p(y)$). Higher is better.

\paragraph{Inception Fr\'echet Inception Distance (iFID)~\cite{xu2026ifid}.}
iFID is a variant of FID designed to be more predictive of generation-stage FID when evaluated at the reconstruction stage. It uses the same formula as rFID but applies a learned feature transformation that better correlates reconstruction quality with downstream generation quality. Lower is better. We report iFID only in the encoder ablation (Table~\ref{tab:rec_encoders}) where it provides additional discriminative power beyond rFID.

\subsection{Downstream Segmentation Metrics}

For segmentation evaluation, we train a 3D~U-Net on full-modality BraTS data (all four real modalities) and evaluate it on inputs where missing modalities have been replaced by each imputation method's output. The segmentation model is never retrained or fine-tuned on imputed data. We evaluate on three tumor sub-regions defined by the BraTS protocol: whole tumor (WT), tumor core (TC), and enhancing tumor (ET), and report the average across all three.

\paragraph{Dice Similarity Coefficient (Dice).}
Dice measures volumetric overlap between the predicted segmentation~$\hat{S}$ and ground-truth segmentation~$S$:
\begin{equation}
    \text{Dice}(\hat{S}, S) = \frac{2|\hat{S} \cap S|}{|\hat{S}| + |S|}
\end{equation}
Dice ranges in $[0, 1]$; higher is better. A Dice of $1$ indicates perfect overlap. We report the mean Dice across the three tumor sub-regions (WT, TC, ET), averaged over all test volumes.

\paragraph{95th-Percentile Hausdorff Distance (HD95).}
HD95 measures the boundary accuracy of the predicted segmentation by computing the 95th percentile of the symmetric surface distances:
\begin{equation}
    \text{HD95}(\hat{S}, S) = \text{percentile}_{95}\!\left(\left\{\min_{s \in \partial S} d(p, s) : p \in \partial \hat{S}\right\} \cup \left\{\min_{p \in \partial \hat{S}} d(s, p) : s \in \partial S\right\}\right)
\end{equation}
where $\partial \hat{S}$ and $\partial S$ denote the surfaces of the predicted and ground-truth segmentations, and $d(\cdot, \cdot)$ is the Euclidean distance. HD95 is reported in millimeters; lower is better. The 95th percentile is used instead of the maximum to reduce sensitivity to outlier voxels. We report the mean HD95 across the three sub-regions, averaged over all test volumes.

\paragraph{Intersection over Union (IoU).}
IoU, also known as the Jaccard index, measures volumetric overlap:
\begin{equation}
    \text{IoU}(\hat{S}, S) = \frac{|\hat{S} \cap S|}{|\hat{S} \cup S|}
\end{equation}
IoU ranges in $[0, 1]$; higher is better. It is related to Dice by $\text{IoU} = \text{Dice}/(2 - \text{Dice})$ and penalizes both false positives and false negatives more heavily than Dice.

\paragraph{Sensitivity (Sens).}
Sensitivity, also known as recall or true positive rate, measures the fraction of ground-truth tumor voxels that are correctly detected:
\begin{equation}
    \text{Sensitivity}(\hat{S}, S) = \frac{|\hat{S} \cap S|}{|S|}
\end{equation}
Sensitivity ranges in $[0, 1]$; higher is better. High sensitivity is clinically important because missed tumor regions (false negatives) can lead to inadequate treatment planning.

\section{Sampling-Step Efficiency}
\label{app:sampling_steps}

\begin{figure}[h]
\centering
\includegraphics[width=0.5\textwidth]{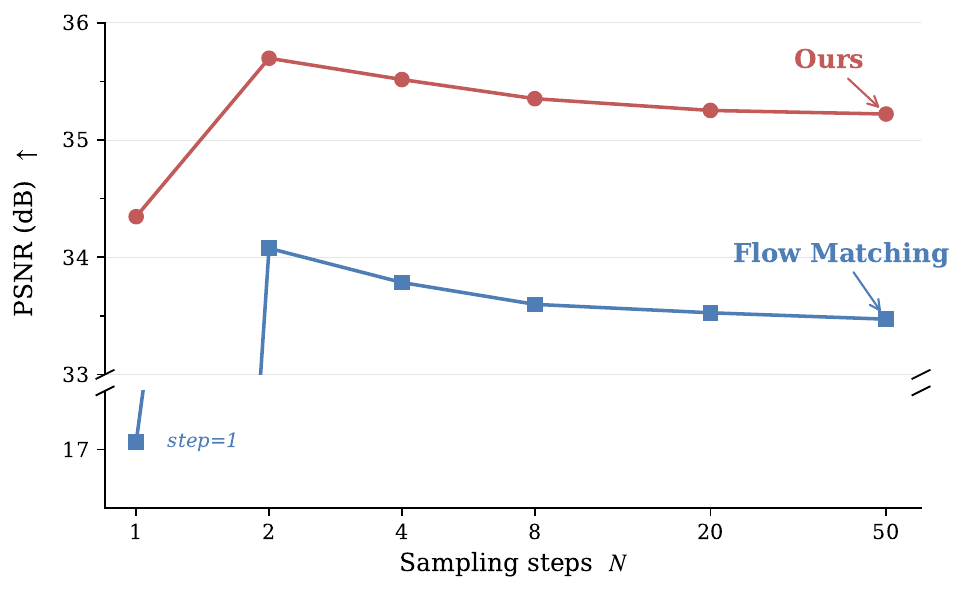}
\caption{PSNR vs.\ Euler steps ($T_1{\to}T_{1c}$). Gaussian FM collapses at $N{=}1$; \method{} starts at $34.3$\,dB. Both saturate by $N{=}2$--$8$; \method{} retains a ${\sim}1.7$\,dB lead.}
\label{fig:psnr_curve}
\end{figure}

Figure~\ref{fig:psnr_curve} compares \method{} against Gaussian-source flow matching (same backbone and tokenizer) as a function of Euler steps~$N$ on $T_1{\to}T_{1c}$. At $N{=}1$ the Gaussian baseline collapses to ${\sim}17$\,dB---one step cannot bridge the $\Theta(\sqrt{d})$ off-manifold gap---while \method{} already reaches $34.3$\,dB. Both saturate by $N{=}2$--$8$; \method{} retains a persistent ${\sim}1.7$\,dB lead, confirming the advantage is structural.

\section{Full N-to-N Synthesis Results}
\label{app:n2n}

This section provides per-combination breakdowns for N-to-N synthesis on BraTS, complementing the averaged results in Table~\ref{tab:n2n_seg_summary} (left).

\begin{table}[h]
    \centering
    \caption{Full N-to-N synthesis results on BraTS (all 14 input combinations).}
    \label{app:tab_n2n}
    \resizebox{\textwidth}{!}{
    \begin{tabular}{cccclccccccccccccccccc}
  \multicolumn{4}{c}{Modalities}                & \multicolumn{1}{c}{} & \multicolumn{8}{c}{PSNR $\uparrow$}                                                                       &                      & \multicolumn{8}{c}{SSIM $\uparrow$}                                                                               \\
  $T_1$     & $T_2$     & $T_{1c}$  & $T_f$     &                      & MMSyn & MMGAN  & MMT    & ModImp & M2DN   & Zhang          & CodeBrain   & \textbf{Ours} & \multicolumn{1}{l}{} & MMSyn      & MMGAN  & MMT    & ModImp & M2DN   & Zhang           & CodeBrain     & \textbf{Ours} \\ \hline
  $\bullet$ & $\circ$   & $\circ$   & $\circ$   &                      & 30.52 & 31.51  & 31.65  & 31.33  & 30.53  & {\ul 33.54}    & 32.99       & \textbf{34.52}      & \multicolumn{1}{l}{} & 0.9130     & 0.9142 & 0.9076 & 0.9169 & 0.8818 & {\ul 0.9393}    & 0.9384        & \textbf{0.9444}     \\
  $\circ$   & $\bullet$ & $\circ$   & $\circ$   &                      & 29.03 & 31.38  & 31.49  & 28.88  & 30.10  & {\ul 34.20}    & 30.93       & \textbf{35.26}      & \multicolumn{1}{l}{} & 0.9003     & 0.9161 & 0.9015 & 0.8978 & 0.8792 & {\ul 0.9441}    & 0.9222        & \textbf{0.9498}     \\
  $\circ$   & $\circ$   & $\bullet$ & $\circ$   &                      & 32.29 & 34.29  & 32.21  & 32.07  & 31.99  & {\ul 34.87}    & 34.05       & \textbf{36.01}      & \multicolumn{1}{l}{} & 0.9196     & 0.9342 & 0.9083 & 0.9193 & 0.8907 & {\ul 0.9454}    & 0.9430        & \textbf{0.9509}     \\
  $\circ$   & $\circ$   & $\circ$   & $\bullet$ &                      & 28.45 & 30.47  & 30.38  & 30.04  & 29.48  & {\ul 32.28}    & 31.39       & \textbf{33.51}      & \multicolumn{1}{l}{} & 0.8959     & 0.9147 & 0.9003 & 0.9013 & 0.8778 & {\ul 0.9336}    & 0.9302        & \textbf{0.9414}     \\
  $\circ$   & $\circ$   & $\bullet$ & $\bullet$ &                      & 32.23 & 34.63  & 32.28  & 32.48  & 32.19  & {\ul 35.72}    & 34.18       & \textbf{36.35}      & \multicolumn{1}{l}{} & 0.9245     & 0.9407 & 0.9124 & 0.9259 & 0.8947 & {\ul 0.9547}    & 0.9475        & \textbf{0.9569}     \\
  $\circ$   & $\bullet$ & $\bullet$ & $\circ$   &                      & 34.82 & 36.99  & 34.79  & 34.86  & 34.38  & {\ul 38.11}    & 37.01       & \textbf{38.86}      & \multicolumn{1}{l}{} & 0.9404     & 0.9506 & 0.9261 & 0.9395 & 0.9091 & {\ul 0.9632}    & 0.9594        & \textbf{0.9647}     \\
  $\bullet$ & $\bullet$ & $\circ$   & $\circ$   &                      & 32.56 & 34.87  & 33.46  & 33.24  & 32.59  & {\ul 35.82}    & 35.11       & \textbf{36.60}      & \multicolumn{1}{l}{} & 0.9256     & 0.9389 & 0.9199 & 0.9295 & 0.8984 & {\ul 0.9535}    & 0.9499        & \textbf{0.9550}     \\
  $\circ$   & $\bullet$ & $\circ$   & $\bullet$ &                      & 30.84 & 32.63  & 31.59  & 31.69  & 30.15  & {\ul 33.90}    & 32.71       & \textbf{34.75}      & \multicolumn{1}{l}{} & 0.9197     & 0.9314 & 0.9115 & 0.9234 & 0.8898 & {\ul 0.9478}    & 0.9438        & \textbf{0.9507}     \\
  $\bullet$ & $\circ$   & $\bullet$ & $\circ$   &                      & 31.84 & 33.76  & 32.41  & 31.88  & 31.80  & {\ul 34.80}    & 33.49       & \textbf{35.60}      & \multicolumn{1}{l}{} & 0.9183     & 0.9305 & 0.9093 & 0.9176 & 0.8878 & {\ul 0.9475}    & 0.9418        & \textbf{0.9501}     \\
  $\bullet$ & $\circ$   & $\circ$   & $\bullet$ &                      & 30.50 & 32.05  & 30.66  & 31.18  & 30.06  & {\ul 33.32}    & 32.65       & \textbf{34.07}      & \multicolumn{1}{l}{} & 0.9159     & 0.9244 & 0.9043 & 0.9184 & 0.8828 & {\ul 0.9443}    & 0.9411        & \textbf{0.9475}     \\
  $\bullet$ & $\bullet$ & $\bullet$ & $\circ$   &                      & 35.26 & 37.38  & 35.87  & 35.49  & 35.52  & {\ul 38.62}    & 37.47       & \textbf{39.20}      & \multicolumn{1}{l}{} & 0.9390     & 0.9505 & 0.9291 & 0.9360 & 0.9108 & {\ul 0.9638}    & 0.9585        & \textbf{0.9639}     \\
  $\circ$   & $\bullet$ & $\bullet$ & $\bullet$ &                      & 35.36 & 37.49  & 34.76  & 35.22  & 34.26  & {\ul 38.35}    & 36.93       & \textbf{38.80}      & \multicolumn{1}{l}{} & 0.9474     & 0.9572 & 0.9314 & 0.9488 & 0.9135 & {\ul 0.9668}    & 0.9628        & \textbf{0.9670}     \\
  $\bullet$ & $\bullet$ & $\circ$   & $\bullet$ &                      & 31.41 & 33.18  & 31.33  & 32.08  & 30.15  & {\ul 34.08}    & 33.28       & \textbf{34.40}      & \multicolumn{1}{l}{} & 0.9229     & 0.9370 & 0.9129 & 0.9288 & 0.9134 & \textbf{0.9501} & 0.9461        & {\ul 0.9486}        \\
  $\bullet$ & $\circ$   & $\bullet$ & $\bullet$ &                      & 30.27 & 32.68  & 30.72  & 30.75  & 30.38  & {\ul 34.15}    & 32.05       & \textbf{34.35}      & \multicolumn{1}{l}{} & 0.9153     & 0.9340 & 0.9010 & 0.9138 & 0.8801 & \textbf{0.9511} & 0.9398        & {\ul 0.9509}        \\ \hline
  \multicolumn{4}{c}{Average}                   & \multicolumn{1}{c}{} & 31.81 & 33.81  & 32.40  & 32.23  & 31.68  & {\ul 35.13}    & 33.88       & \textbf{35.88}      &                      & 0.9213     & 0.9339 & 0.9125 & 0.9226 & 0.8936 & {\ul 0.9504}    & 0.9446        & \textbf{0.9530}     \\
            &           &           &           &                      &       &        &        &        &        &                &             &                     &                      &            &        &        &        &        &                 &               &                     \\
  \multicolumn{4}{c}{Modalities}                & \multicolumn{1}{c}{} & \multicolumn{8}{c}{rFID $\downarrow$}                                                                       &                      & \multicolumn{8}{c}{IS $\uparrow$}                                                                                 \\
  $T_1$     & $T_2$     & $T_{1c}$  & $T_f$     &                      & MMSyn & MMGAN  & MMT    & ModImp & M2DN   & Zhang          & CodeBrain   & \textbf{Ours} &                      & MMSyn      & MMGAN  & MMT    & ModImp & M2DN   & Zhang           & CodeBrain     & \textbf{Ours} \\ \hline
  $\bullet$ & $\circ$   & $\circ$   & $\circ$   &                      & 43.42 & 120.43 & 114.33 & 136.24 & 107.74 & {\ul 24.43}    & 37.88       & \textbf{19.36}      &                      & 2.58       & 2.43   & 2.38   & 2.34   & 1.98   & \textbf{2.71}   & 2.61          & {\ul 2.69}          \\
  $\circ$   & $\bullet$ & $\circ$   & $\circ$   &                      & 53.02 & 142.49 & 183.83 & 165.24 & 104.92 & {\ul 24.65}    & 41.55       & \textbf{21.56}      &                      & 2.46       & 2.82   & 2.86   & 2.32   & 2.02   & \textbf{3.05}   & 2.89          & {\ul 3.02}          \\
  $\circ$   & $\circ$   & $\bullet$ & $\circ$   &                      & 46.61 & 112.22 & 150.69 & 142.93 & 116.82 & 31.18          & {\ul 30.81} & \textbf{18.63}      &                      & 2.59       & 2.55   & 2.42   & 2.32   & 2.31   & {\ul 2.62}      & 2.46          & \textbf{2.70}       \\
  $\circ$   & $\circ$   & $\circ$   & $\bullet$ &                      & 51.26 & 138.05 & 136.82 & 179.83 & 116.80 & {\ul 39.71}    & 63.41       & \textbf{20.23}      &                      & {\ul 2.75} & 2.44   & 2.56   & 2.20   & 2.13   & 2.65            & 2.43          & \textbf{2.83}       \\
  $\circ$   & $\circ$   & $\bullet$ & $\bullet$ &                      & 37.64 & 85.16  & 138.45 & 112.86 & 98.33  & {\ul 18.62}    & 24.93       & \textbf{18.45}      &                      & 2.23       & 2.35   & 2.34   & 2.37   & 1.99   & {\ul 2.46}      & 2.28          & \textbf{2.62}       \\
  $\circ$   & $\bullet$ & $\bullet$ & $\circ$   &                      & 23.42 & 77.00  & 159.32 & 106.32 & 97.89  & \textbf{13.83} & {\ul 17.30} & 19.09               &                      & 2.91       & 2.75   & 2.73   & 2.50   & 2.48   & {\ul 2.93}      & 2.91          & \textbf{2.98}       \\
  $\bullet$ & $\bullet$ & $\circ$   & $\circ$   &                      & 42.03 & 78.23  & 123.54 & 117.95 & 107.37 & \textbf{17.73} & 36.79       & {\ul 18.66}         &                      & 2.89       & 2.65   & 2.62   & 2.48   & 2.37   & \textbf{2.98}   & 2.92          & {\ul 2.95}          \\
  $\circ$   & $\bullet$ & $\circ$   & $\bullet$ &                      & 39.17 & 109.05 & 146.10 & 122.82 & 115.78 & \textbf{23.53} & 37.00       & {\ul 23.72}         &                      & 3.20       & 2.77   & 2.81   & 2.70   & 2.61   & \textbf{3.21}   & 3.07          & {\ul 3.16}          \\
  $\bullet$ & $\circ$   & $\bullet$ & $\circ$   &                      & 30.91 & 89.42  & 121.55 & 130.63 & 89.65  & \textbf{17.70} & 25.21       & {\ul 18.90}         &                      & 2.30       & 2.34   & 2.19   & 2.22   & 1.84   & \textbf{2.60}   & 2.44          & {\ul 2.49}          \\
  $\bullet$ & $\circ$   & $\circ$   & $\bullet$ &                      & 37.53 & 86.17  & 106.29 & 118.49 & 107.84 & \textbf{15.68} & 39.22       & {\ul 19.10}         &                      & {\ul 2.49} & 2.10   & 2.19   & 2.39   & 1.87   & 2.43            & 2.22          & \textbf{2.55}       \\
  $\bullet$ & $\bullet$ & $\bullet$ & $\circ$   &                      & 24.88 & 72.50  & 135.83 & 133.04 & 84.98  & \textbf{12.35} & 19.12       & {\ul 16.73}         &                      & 2.91       & 2.72   & 2.61   & 2.37   & 2.37   & \textbf{2.99}   & \textbf{2.99} & 2.83                \\
  $\circ$   & $\bullet$ & $\bullet$ & $\bullet$ &                      & 19.03 & 55.59  & 170.43 & 65.03  & 100.85 & \textbf{9.89}  & {\ul 11.58} & 22.39               &                      & 3.06       & 2.69   & 2.63   & 2.78   & 2.72   & {\ul 3.08}      & 3.03          & \textbf{3.16}       \\
  $\bullet$ & $\bullet$ & $\circ$   & $\bullet$ &                      & 49.04 & 63.72  & 113.17 & 85.78  & 124.11 & \textbf{20.45} & 48.63       & {\ul 21.04}         &                      & 2.89       & 2.61   & 2.42   & 2.88   & 2.46   & {\ul 2.96}      & 2.89          & \textbf{3.07}       \\
  $\bullet$ & $\circ$   & $\bullet$ & $\bullet$ &                      & 21.65 & 73.67  & 100.38 & 116.96 & 85.34  & \textbf{10.04} & 20.84       & {\ul 17.62}         &                      & 1.75       & 1.88   & 1.93   & 2.02   & 1.30   & {\ul 2.05}      & 1.70          & \textbf{2.08}       \\ \hline
  \multicolumn{4}{c}{Average}                   & \multicolumn{1}{c}{} & 37.12 & 93.12  & 135.77 & 123.87 & 104.17 & {\ul 19.98}    & 32.45       & \textbf{19.68}      &                      & 2.64       & 2.51   & 2.48   & 2.42   & 2.18   & {\ul 2.76}      & 2.63          & \textbf{2.80}
  \end{tabular}
  }
\end{table}

\paragraph{Analysis by number of available sources.}
A clear trend emerges when the 14~combinations are grouped by how many modalities are observed.

\emph{Single-source (rows 1--4).}
This is the hardest regime: three out of four modalities must be hallucinated from a single observation. \method{} achieves the best PSNR on all four single-source settings, with gains of $+0.98$\,dB (from $T_1$), $+1.06$\,dB (from $T_2$), $+1.14$\,dB (from $T_{1c}$), and $+1.23$\,dB (from $T_f$) over the runner-up Zhang. In rFID the advantage is even more pronounced: the single-source average rFID for \method{} is $19.95$, versus $29.99$ for Zhang ($-33.5\%$). This confirms that when only one modality is available, the short transport distance on the hypersphere matters most---the generator needs to cover three missing directions, and starting from an on-manifold semantic feature rather than noise provides a much stronger prior.

Notably, the FLAIR-only setting ($T_f$-only) is the weakest single source for all baselines (e.g., Zhang drops to $32.28$\,dB PSNR, the lowest among single-source rows), yet \method{} still reaches $33.51$\,dB---demonstrating robustness even when the source modality carries the least contrast-specific information.

\emph{Two-source (rows 5--10).}
With two observed modalities, latent averaging places the source representation closer to every target, and all methods improve. \method{} leads in PSNR on all six two-source settings and achieves the best rFID on three of six. On the remaining three, Zhang holds a small rFID edge (e.g., $13.83$ vs.\ $19.09$ for $\{T_2, T_{1c}\}$). This pattern---\method{} dominating pixel-level fidelity while occasionally trailing on distributional metrics in multi-source regimes---is consistent with Zhang's VAE decoder producing slightly sharper textures when ample source information is available, whereas \method{}'s DINO-based decoder prioritizes semantic consistency.

\emph{Three-source (rows 11--14).}
When three modalities are observed, only one must be imputed, and the task becomes substantially easier for all methods. Performance differences narrow: \method{} leads on PSNR in three of four settings and ties or slightly trails Zhang on SSIM in two. The average three-source PSNR for \method{} is $36.69$\,dB versus $36.30$\,dB for Zhang ($+0.39$\,dB), a smaller gap than in the single-source regime ($+1.10$\,dB). This diminishing marginal gain is expected: with three sources, even simple averaging produces a high-quality latent near the target, so the geometric advantage of on-manifold transport is less decisive.

\paragraph{Per-metric observations.}
\emph{PSNR and SSIM} are strongly correlated across all settings, and \method{} achieves the best average on both ($35.88$\,dB, $0.9530$). The SSIM advantage is most visible in single-source regimes where structural preservation matters most.

\emph{rFID.} \method{} achieves the best overall average ($19.68$ vs.\ $19.98$ for Zhang), but the per-combination picture is nuanced: \method{} dominates in single-source rFID (4/4 best) while Zhang leads in several multi-source settings. We attribute this to the fact that Zhang's per-modality VAE decoder can leverage multiple source cues to sharpen textures, whereas \method{}'s decoder operates on a single averaged latent. Despite this, \method{}'s average rFID remains the best across all 14 combinations.

\emph{IS.} \method{} achieves the highest average IS ($2.80$), indicating that synthesized images are both diverse and recognizable. The IS advantage is largest in single-source settings, consistent with the PSNR and rFID trends.

\section{Full Segmentation Results}
\label{app:seg}

This section provides per-combination breakdowns for downstream brain tumor segmentation on BraTS, complementing the averaged results in Table~\ref{tab:n2n_seg_summary} (right). The segmentation model is a 3D~U-Net trained \emph{once} on full-modality (all four real modalities) BraTS data. At test time, missing modalities are filled in by each imputation method; the segmentation model is never retrained or fine-tuned. This protocol isolates the quality of the imputed modalities: better imputation translates directly to higher segmentation accuracy.

\begin{table}[h]
    \centering
    \caption{Full downstream segmentation results on BraTS (all 14 input combinations).}
    \label{app:tab_seg}
    \resizebox{\textwidth}{!}{
    \begin{tabular}{cccclccccccccccccccccccc}
  \multicolumn{4}{c}{Modalities}                                                            & \multicolumn{1}{c}{} & \multicolumn{9}{c}{Dice}                                                                                                                                                                       & \multicolumn{1}{l}{} & \multicolumn{9}{c}{HD95}                                                                                                                                                                       \\
  $T_1$                & $T_2$                & $T_{1c}$             & $T_f$                &                      & Zero   & MMSyn                & MMGAN                & MMT                  & ModImp               & M2DN                 & Zhang                & CodeBrain            & \textbf{Ours}  &                      & Zero                 & MMSyn  & MMGAN                & MMT                  & ModImp               & M2DN                 & Zhang                & CodeBrain            & \textbf{Ours}  \\ \hline
  $\bullet$            & $\circ$              & $\circ$              & $\circ$              &                      & 0.0094 & 0.0686               & 0.1852               & 0.1910               & 0.0851               & 0.1558               & 0.1570               & {\ul 0.2094}         & \textbf{0.5710}      &                      & 316.87               & 237.74 & 167.35               & {\ul 94.73}          & 232.55               & 140.21               & 114.11               & 169.75               & \textbf{35.75}       \\
  $\circ$              & $\bullet$            & $\circ$              & $\circ$              &                      & 0.0133 & 0.3025               & 0.2740               & 0.2696               & 0.2679               & 0.1796               & 0.2662               & {\ul 0.3637}         & \textbf{0.6477}      &                      & 326.12               & {\ul 65.63}          & 94.85                & 94.31                & 77.71                & 131.71               & 71.67                & 91.35                & \textbf{23.15}       \\
  $\circ$              & $\circ$              & $\bullet$            & $\circ$              &                      & 0.0257 & 0.6412               & 0.6029               & 0.5389               & 0.6125               & 0.5467               & 0.3932               & {\ul 0.6771}         & \textbf{0.7811}      &                      & 301.14               & {\ul 35.79}          & 66.93                & 58.71                & 43.21                & 62.99                & 75.44                & 38.97                & \textbf{22.09}       \\
  $\circ$              & $\circ$              & $\circ$              & $\bullet$            &                      & 0.3695 & 0.4096               & 0.3583               & 0.3395               & {\ul 0.4633}         & 0.3508               & 0.4260               & 0.3946               & \textbf{0.6574}      &                      & 105.40               & 68.57  & 75.06                & 62.55                & 55.85                & 77.51                & {\ul 51.32}          & 97.50                & \textbf{22.16}       \\
  $\circ$              & $\circ$              & $\bullet$            & $\bullet$            &                      & 0.7386 & 0.8284               & 0.8348               & 0.8173               & 0.8254               & 0.7897               & 0.8195               & {\ul 0.8421}         & \textbf{0.8512}      &                      & 28.89                & 13.31  & {\ul 13.07}          & 17.19                & 14.66                & 20.98                & 14.49                & 13.71                & \textbf{12.05}       \\
  $\circ$              & $\bullet$            & $\bullet$            & $\circ$              &                      & 0.0326 & 0.7345               & 0.7771               & 0.7497               & 0.7715               & 0.7488               & 0.5505               & {\ul 0.8088}         & \textbf{0.8360}      &                      & 304.90               & 28.07  & 20.52                & 22.61                & 20.58                & 23.56                & 55.67                & {\ul 17.48}          & \textbf{14.21}       \\
  $\bullet$            & $\bullet$            & $\circ$              & $\circ$              &                      & 0.0645 & 0.2702               & 0.3634               & 0.3178               & 0.3285               & 0.3423               & 0.3321               & {\ul 0.3752}         & \textbf{0.6706}      &                      & 260.65               & 90.64  & 66.87                & 69.16                & 105.50               & {\ul 61.82}          & 63.14                & 82.26                & \textbf{22.89}       \\
  $\circ$              & $\bullet$            & $\circ$              & $\bullet$            &                      & 0.4362 & 0.4498               & 0.4463               & 0.3922               & 0.4542               & 0.4578               & {\ul 0.4962}         & 0.4452               & \textbf{0.6991}      &                      & 84.39                & 56.88  & 55.82                & 53.63                & 51.04                & {\ul 42.15}          & 47.37                & 71.33                & \textbf{19.43}       \\
  $\bullet$            & $\circ$              & $\bullet$            & $\circ$              &                      & 0.0147 & 0.7196               & 0.7021               & 0.6940               & 0.7007               & 0.6350               & 0.3675               & {\ul 0.7439}         & \textbf{0.7991}      &                      & 345.43               & {\ul 26.04}          & 40.16                & 32.35                & 31.11                & 43.96                & 105.77               & 29.84                & \textbf{20.33}       \\
  $\bullet$            & $\circ$              & $\circ$              & $\bullet$            &                      & 0.4143 & 0.4325               & 0.4298               & 0.4109               & 0.4348               & 0.4107               & {\ul 0.4581}         & 0.4532               & \textbf{0.6930}      &                      & 81.95                & 71.79  & 63.10                & 51.80                & 75.35                & 56.53                & {\ul 43.54}          & 73.44                & \textbf{19.01}       \\
  $\bullet$            & $\bullet$            & $\bullet$            & $\circ$              &                      & 0.2744 & 0.7687               & 0.8018               & 0.7872               & 0.7955               & 0.7891               & 0.6800               & {\ul 0.8130}         & \textbf{0.8411}      &                      & 126.01               & 22.48  & 22.11                & 19.47                & 18.75                & 18.62                & 33.03                & {\ul 18.58}          & \textbf{13.67}       \\
  $\circ$              & $\bullet$            & $\bullet$            & $\bullet$            &                      & 0.7975 & 0.8392               & {\ul 0.8518}         & 0.8413               & 0.8462               & 0.8440               & 0.8215               & 0.8501               & \textbf{0.8535}      &                      & 16.35                & {\ul 11.18}          & 11.63                & 12.83                & 12.13                & 12.49                & 14.50                & \textbf{11.04}       & 11.34                \\
  $\bullet$            & $\bullet$            & $\circ$              & $\bullet$            &                      & 0.4468 & 0.4553               & 0.4433               & 0.4353               & 0.4456               & 0.4920               & {\ul 0.5016}         & 0.4738               & \textbf{0.7085}      &                      & 60.15                & 67.39  & 65.76                & 45.09                & 81.65                & {\ul 37.92}          & 40.47                & 62.34                & \textbf{17.68}       \\
  $\bullet$            & $\circ$              & $\bullet$            & $\bullet$            &                      & 0.8214 & 0.8441               & 0.8484               & 0.8453               & 0.8462               & 0.8085               & 0.8428               & {\ul 0.8520}         & \textbf{0.8533}      &                      & 13.40                & 12.71  & 12.25                & 13.30                & {\ul 12.05}          & 16.11                & 12.33                & \textbf{11.63}       & 12.51                \\ \hline
  \multicolumn{4}{c}{Average}                                                               & \multicolumn{1}{c}{} & 0.3185 & 0.5546               & 0.5657               & 0.5450               & 0.5627               & 0.5393               & 0.5080               & {\ul 0.5930}         & \textbf{0.7473}      &                      & 169.40               & 57.73  & 55.39                & {\ul 46.27}          & 59.44                & 53.32                & 53.06                & 56.37                & \textbf{19.02}       \\ \hline
  \multicolumn{4}{c}{Full}                                                                  &                      & \multicolumn{9}{c}{0.8554}                                                                                                                                                                     & \multicolumn{1}{l}{} & \multicolumn{9}{c}{11.41}                                                                                                                                                                      \\
  \multicolumn{1}{l}{} & \multicolumn{1}{l}{} & \multicolumn{1}{l}{} & \multicolumn{1}{l}{} & \multicolumn{1}{c}{} &        & \multicolumn{1}{l}{} & \multicolumn{1}{l}{} & \multicolumn{1}{l}{} & \multicolumn{1}{l}{} & \multicolumn{1}{l}{} & \multicolumn{1}{l}{} & \multicolumn{1}{l}{} & \multicolumn{1}{l}{} & \multicolumn{1}{l}{} &        & \multicolumn{1}{l}{} & \multicolumn{1}{l}{} & \multicolumn{1}{l}{} & \multicolumn{1}{l}{} & \multicolumn{1}{l}{} & \multicolumn{1}{l}{} & \multicolumn{1}{l}{} & \multicolumn{1}{l}{} \\
  \multicolumn{4}{c}{Modalities}                                                            & \multicolumn{1}{c}{} & \multicolumn{9}{c}{IoU}                                                                                                                                                                        & \multicolumn{1}{l}{} & \multicolumn{9}{c}{Sensitivity}                                                                                                                                                                \\
  $T_1$                & $T_2$                & $T_{1c}$             & $T_f$                &                      & Zero   & MMSyn                & MMGAN                & MMT                  & ModImp               & M2DN                 & Zhang                & CodeBrain            & \textbf{Ours}  &                      & Zero                 & MMSyn  & MMGAN                & MMT                  & ModImp               & M2DN                 & Zhang                & CodeBrain            & \textbf{Ours}  \\ \hline
  $\bullet$            & $\circ$              & $\circ$              & $\circ$              &                      & 0.0094 & 0.0441               & 0.1308               & 0.1297               & 0.0576               & 0.1002               & 0.1044               & {\ul 0.1547}         & \textbf{0.4489}      &                      & 0.0094               & 0.0792 & 0.1498               & {\ul 0.1867}         & 0.0689               & 0.1185               & 0.1300               & 0.1764               & \textbf{0.5651}      \\
  $\circ$              & $\bullet$            & $\circ$              & $\circ$              &                      & 0.0116 & 0.2138               & 0.2012               & 0.2013               & 0.1945               & 0.1323               & 0.1884               & {\ul 0.2898}         & \textbf{0.5310}      &                      & 0.0130               & {\ul 0.4367}         & 0.2637               & 0.2683               & 0.2668               & 0.1545               & 0.2410               & 0.3458               & \textbf{0.6551}      \\
  $\circ$              & $\circ$              & $\bullet$            & $\circ$              &                      & 0.0183 & 0.5381               & 0.5070               & 0.4369               & 0.5006               & 0.4413               & 0.2965               & {\ul 0.5835}         & \textbf{0.6891}      &                      & 0.0283               & 0.6090 & 0.5446               & 0.4974               & 0.5997               & 0.4753               & 0.3478               & {\ul 0.6518}         & \textbf{0.7663}      \\
  $\circ$              & $\circ$              & $\circ$              & $\bullet$            &                      & 0.2878 & 0.1931               & 0.2828               & 0.2643               & {\ul 0.3710}         & 0.2543               & 0.3335               & 0.3314               & \textbf{0.5468}      &                      & 0.4902               & 0.4211 & 0.3789               & 0.3566               & {\ul 0.5050}         & 0.2921               & 0.4308               & 0.3741               & \textbf{0.6585}      \\
  $\circ$              & $\circ$              & $\bullet$            & $\bullet$            &                      & 0.6413 & 0.7461               & 0.7562               & 0.7330               & 0.7398               & 0.6929               & 0.7344               & {\ul 0.7651}         & \textbf{0.7768}      &                      & 0.7033               & 0.8246 & 0.8222               & 0.8103               & 0.8242               & 0.7460               & 0.8065               & {\ul 0.8452}         & \textbf{0.8496}      \\
  $\circ$              & $\bullet$            & $\bullet$            & $\circ$              &                      & 0.0237 & 0.6350               & 0.6857               & 0.6518               & 0.6731               & 0.6521               & 0.4519               & {\ul 0.7228}         & \textbf{0.7578}      &                      & 0.0238               & 0.7145 & 0.7427               & 0.7307               & 0.7601               & 0.7081               & 0.5071               & {\ul 0.7995}         & \textbf{0.8426}      \\
  $\bullet$            & $\bullet$            & $\circ$              & $\circ$              &                      & 0.0442 & 0.2016               & 0.2893               & 0.2418               & 0.2581               & 0.2551               & 0.2445               & {\ul 0.3028}         & \textbf{0.5555}      &                      & 0.0460               & 0.2838 & {\ul 0.3446}         & 0.3015               & 0.3122               & 0.3167               & 0.2901               & 0.3452               & \textbf{0.6762}      \\
  $\circ$              & $\bullet$            & $\circ$              & $\bullet$            &                      & 0.3614 & 0.3651               & 0.3667               & 0.3191               & 0.3774               & 0.3546               & {\ul 0.3988}         & 0.3774               & \textbf{0.5905}      &                      & 0.4828               & 0.4425 & 0.4805               & 0.3889               & 0.4568               & 0.4168               & {\ul 0.5029}         & 0.4237               & \textbf{0.7017}      \\
  $\bullet$            & $\circ$              & $\bullet$            & $\circ$              &                      & 0.0127 & 0.6239               & 0.6070               & 0.5916               & 0.6033               & 0.5279               & 0.2957               & {\ul 0.6510}         & \textbf{0.7116}      &                      & 0.0127               & 0.6874 & 0.6607               & 0.6725               & 0.6776               & 0.5741               & 0.3326               & {\ul 0.7184}         & \textbf{0.7926}      \\
  $\bullet$            & $\circ$              & $\circ$              & $\bullet$            &                      & 0.3396 & 0.3551               & 0.3572               & 0.3310               & 0.3637               & 0.3126               & 0.3684               & {\ul 0.3823}         & \textbf{0.5833}      &                      & {\ul 0.4638}         & 0.4062 & 0.4244               & 0.4191               & 0.4207               & 0.3758               & 0.4487               & 0.4261               & \textbf{0.6904}      \\
  $\bullet$            & $\bullet$            & $\bullet$            & $\circ$              &                      & 0.2050 & 0.6752               & 0.7175               & 0.6950               & 0.7082               & 0.6958               & 0.5775               & {\ul 0.7302}         & \textbf{0.7631}      &                      & 0.2118               & 0.7548 & 0.7895               & 0.7756               & 0.7845               & 0.7644               & 0.6291               & {\ul 0.8041}         & \textbf{0.8474}      \\
  $\circ$              & $\bullet$            & $\bullet$            & $\bullet$            &                      & 0.7085 & 0.7609               & {\ul 0.7781}         & 0.7654               & 0.7702               & 0.7700               & 0.7392               & 0.7760               & \textbf{0.7811}      &                      & 0.8118               & 0.8328 & 0.8479               & 0.8365               & 0.8481               & 0.8391               & 0.8344               & {\ul 0.8500}         & \textbf{0.8546}      \\
  $\bullet$            & $\bullet$            & $\circ$              & $\bullet$            &                      & 0.3698 & 0.3772               & 0.3753               & 0.3575               & 0.3786               & 0.3970               & {\ul 0.4057}         & 0.3996               & \textbf{0.5997}      &                      & 0.4594               & 0.4412 & 0.4363               & 0.4297               & 0.4288               & 0.4772               & {\ul 0.4896}         & 0.4479               & \textbf{0.7057}      \\
  $\bullet$            & $\circ$              & $\bullet$            & $\bullet$            &                      & 0.7373 & 0.7687               & 0.7738               & 0.7683               & 0.7692               & 0.7232               & 0.7641               & {\ul 0.7777}         & \textbf{0.7802}      &                      & 0.8080               & 0.8407 & 0.8476               & 0.8497               & 0.8500               & 0.7806               & 0.8308               & {\ul 0.8548}         & \textbf{0.8556}      \\ \hline
  \multicolumn{4}{c}{Average}                                                               & \multicolumn{1}{c}{} & 0.2693 & 0.4641               & 0.4877               & 0.4633               & 0.4832               & 0.4507               & 0.4216               & {\ul 0.5174}         & \textbf{0.6511}      &                      & 0.3260               & 0.5553 & 0.5524               & 0.5374               & 0.5574               & 0.5028               & 0.4872               & {\ul 0.5759}         & \textbf{0.7418}      \\ \hline
  \multicolumn{4}{c}{Full}                                                                  &                      & \multicolumn{9}{c}{0.7829}                                                                                                                                                                     & \multicolumn{1}{l}{} & \multicolumn{9}{c}{0.8583}
  \end{tabular}
  }
\end{table}

\paragraph{Analysis by number of available sources.}

\emph{Single-source (rows 1--4).}
This is the most challenging and clinically relevant regime: only one modality is observed, and the segmentation model must operate on three imputed channels. All baselines collapse in at least some single-source settings---for example, with $T_1$-only input, the best baseline (CodeBrain) reaches only Dice $0.2094$, far below clinical utility. \method{} dramatically outperforms all baselines in every single-source case: Dice scores of $0.5710$ ($T_1$-only), $0.6477$ ($T_2$-only), $0.7811$ ($T_{1c}$-only), and $0.6574$ ($T_f$-only). The gap over the runner-up CodeBrain ranges from $+0.10$ ($T_{1c}$-only) to $+0.36$ ($T_1$-only) in Dice, and HD95 improves by $3\text{--}5\times$ (e.g., $169.75 \to 35.75$ for $T_1$-only). This result is particularly striking because the segmentation model was never exposed to imputed data during training---\method{}'s synthesized modalities are realistic enough to be consumed directly by a model trained on real data.

Among the four single-source inputs, $T_{1c}$-only yields the highest Dice ($0.7811$) for \method{}, which is expected: contrast-enhanced $T_{1c}$ highlights the tumor core, providing the richest signal for segmentation. Conversely, $T_1$-only is the hardest ($0.5710$), as non-contrast $T_1$ lacks tumor-specific signal. Even so, \method{} with $T_1$-only surpasses every baseline's performance with \emph{any} single source.

\emph{Two-source (rows 5--10).}
With two observed modalities, segmentation accuracy improves across all methods. The key finding is that \method{} remains the best on all six two-source combinations, with particularly large gains when $T_{1c}$ is missing. For example, with $\{T_1, T_2\}$ observed (no $T_{1c}$, no $T_f$), baselines plateau around Dice $0.33$--$0.38$, while \method{} reaches $0.6706$ ($+0.30$ over CodeBrain). This indicates that \method{} successfully synthesizes the contrast-enhanced features that the segmentation model relies on most.

When $T_{1c}$ \emph{is} available as one of the two sources (rows 5, 6, 9), all methods perform well (Dice $> 0.75$), and the gap between \method{} and baselines narrows. This is consistent with the interpretation that the segmentation model's performance is bottlenecked by $T_{1c}$ quality, and when the real $T_{1c}$ is present, imputation quality of other modalities matters less.

\emph{Three-source (rows 11--14).}
With three modalities observed, only one is imputed, and performance approaches the full-modality oracle for all methods. \method{} still leads on 12 out of 14 Dice entries in this regime. The two settings where baselines are competitive are $\{T_2, T_{1c}, T_f\}$ (missing $T_1$) and $\{T_1, T_{1c}, T_f\}$ (missing $T_2$)---both cases where the missing modality is relatively redundant given the remaining three. Here, \method{} achieves Dice $0.8535$ and $0.8533$ respectively, within $0.002$--$0.005$ of the full oracle ($0.8554$), effectively solving the imputation problem.

\paragraph{Gap to the full-modality oracle.}
The full oracle (all four real modalities) achieves Dice $0.8554$ and HD95 $11.41$. \method{} reaches an average Dice of $0.7473$ and HD95 of $19.02$---closing $79.8\%$ of the Dice gap and $95.2\%$ of the HD95 gap relative to the zero-fill baseline (Dice $0.3185$, HD95 $169.40$). By contrast, the best baseline CodeBrain closes only $51.1\%$ of the Dice gap and $71.5\%$ of the HD95 gap. The advantage is most dramatic in HD95: \method{}'s average of $19.02$ is $2.4\times$ better than the best baseline ($46.27$ for MMT) and only $1.7\times$ the oracle, indicating that imputed modalities preserve spatial accuracy of tumor boundaries.

\paragraph{Per-metric observations.}
\emph{IoU} closely tracks Dice across all settings, with \method{} leading on 14/14 combinations. The average IoU of $0.6511$ is $+0.1337$ above the best baseline ($0.5174$ for CodeBrain), a $25.8\%$ relative improvement.

\emph{Sensitivity} measures the fraction of true tumor voxels correctly detected. \method{} achieves the highest average sensitivity ($0.7418$), $+0.1659$ above CodeBrain ($0.5759$). High sensitivity is clinically critical---missed tumor regions can lead to under-treatment. The largest sensitivity gains appear in single-source settings (e.g., $0.5651$ vs.\ $0.1764$ for $T_1$-only), where baselines catastrophically under-detect tumors.

\paragraph{Synthesis quality vs.\ segmentation performance.}
An interesting observation is the disconnect between synthesis fidelity and downstream utility for some baselines. Zhang achieves the second-best synthesis metrics (PSNR, rFID) but ranks only sixth in segmentation Dice ($0.5080$), well below CodeBrain ($0.5930$) and even MMSyn ($0.5546$). This suggests that Zhang's VAE-based decoder produces images that look realistic but do not preserve the fine-grained diagnostic features (e.g., tumor boundaries, contrast enhancement patterns) that the segmentation model expects. \method{} does not exhibit this disconnect: it leads on \emph{both} synthesis fidelity and downstream segmentation, confirming that the DINO-based hyperspherical representation preserves semantically meaningful content through the imputation pipeline.

\section{Qualitative Results}
\label{app:qualitative}

\begin{figure}[h]
\centering
\includegraphics[width=\textwidth]{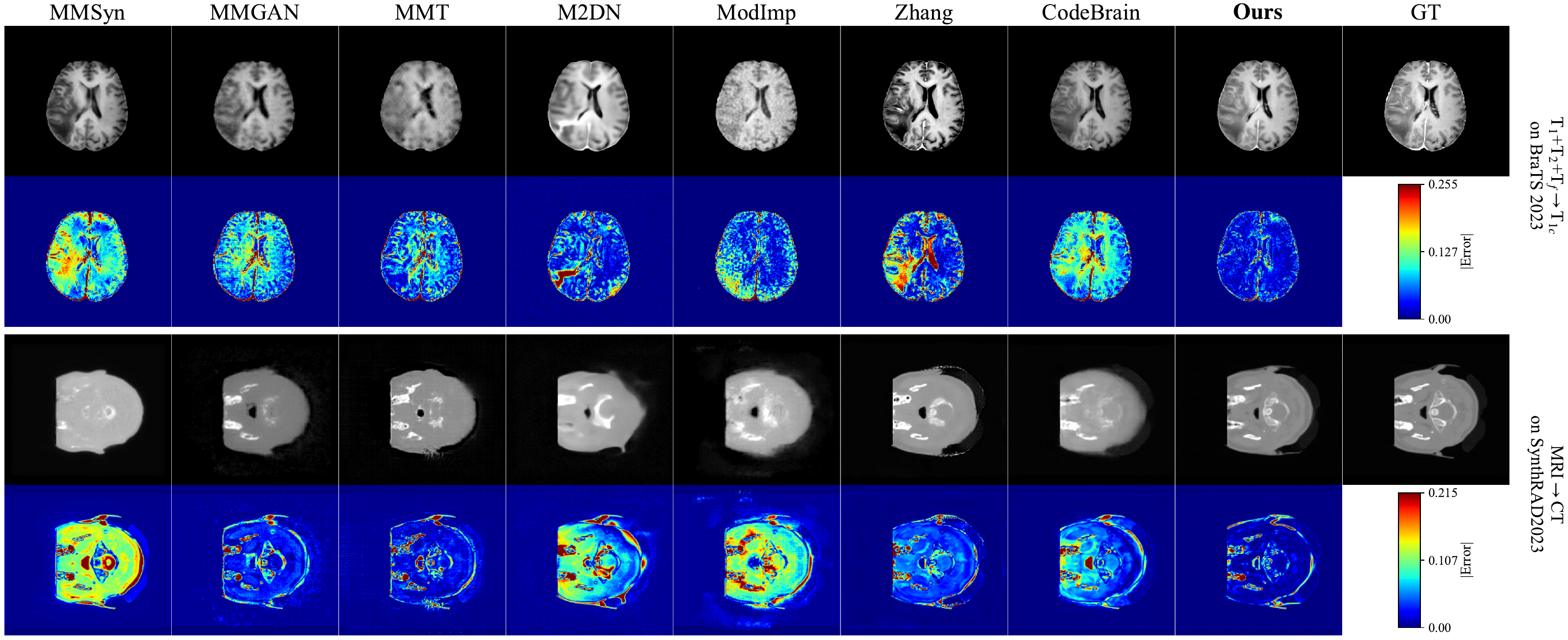}
\caption{\textbf{Qualitative comparison of synthesized images and error maps.} \textbf{Top two rows:} $T_1{+}T_2{+}T_f \to T_{1c}$ on BraTS. Row~1 shows synthesized $T_{1c}$ images; row~2 shows the corresponding per-pixel absolute error (jet colormap, darker blue = lower error). \textbf{Bottom two rows:} $\mathrm{MRI} \to \mathrm{CT}$ on SynthRAD2023, with the same layout. Columns correspond to seven baselines, \method{}, and the ground truth (GT).}
\label{fig:qual_synth}
\end{figure}

\paragraph{Synthesis quality (Figure~\ref{fig:qual_synth}).}
On BraTS $T_1{+}T_2{+}T_f \to T_{1c}$ (top), most baselines produce blurry outputs that lose fine anatomical detail---MMSyn and MMT wash out cortical sulci, while M2DN introduces intensity artifacts in the ventricles. Zhang and CodeBrain recover overall brain structure but exhibit elevated error in the tumor region, where contrast enhancement is hardest to hallucinate from non-contrast inputs. \method{} produces the sharpest reconstruction with the lowest error across both healthy tissue and the tumor. The error map (row~2) confirms this quantitatively: \method{}'s error is uniformly low (predominantly dark blue), whereas all baselines show concentrated high-error patches (red/orange) around the tumor core and cortical boundaries.

On SynthRAD2023 $\mathrm{MRI} \to \mathrm{CT}$ (bottom), the cross-domain gap is larger due to fundamentally different imaging physics. Several baselines (MMSyn, M2DN, ModImp) fail to reproduce bone density contrast, producing washed-out or inverted intensities. Zhang generates plausible CT contrast but with visible ring-like artifacts along the skull boundary. \method{} best reproduces the characteristic CT bone--soft-tissue contrast, and its error map shows the least residual error, particularly along the skull and internal structures where HU accuracy matters most for radiotherapy planning.

\begin{figure}[h]
\centering
\includegraphics[width=\textwidth]{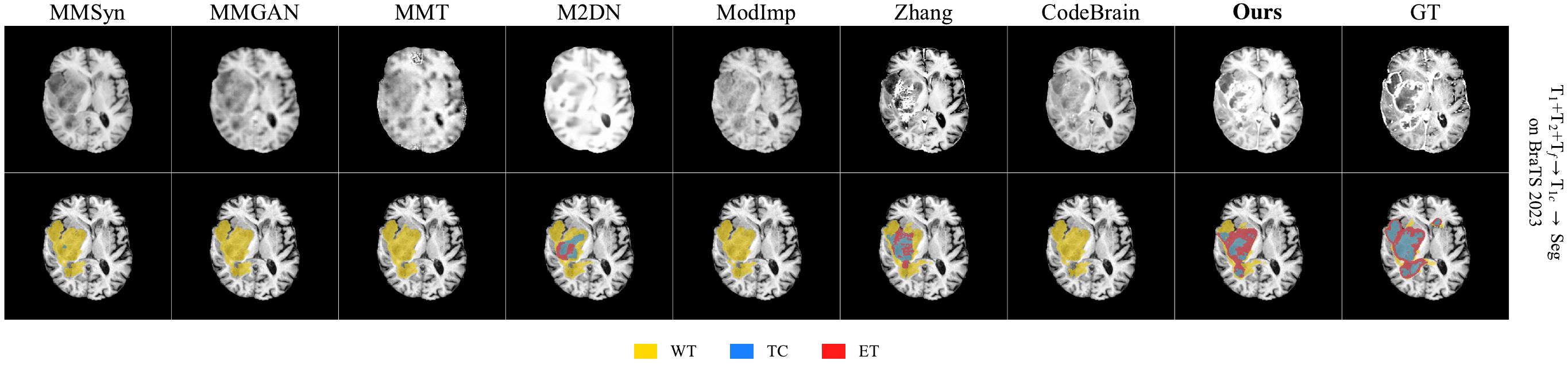}
\caption{\textbf{Qualitative comparison of downstream segmentation.} Setting: $T_1{+}T_2{+}T_f \to T_{1c}$ on BraTS, then feeding all four modalities (three real + one imputed) to a 3D~U-Net trained on full-modality data. Row~1: synthesized $T_{1c}$ images. Row~2: segmentation overlays showing whole tumor (WT, yellow), tumor core (TC, blue), and enhancing tumor (ET, red). The rightmost column is the ground truth.}
\label{fig:qual_seg}
\end{figure}

\paragraph{Downstream segmentation (Figure~\ref{fig:qual_seg}).}
The segmentation overlays reveal why synthesis fidelity translates (or fails to translate) into downstream accuracy. Most baselines produce segmentations that capture the rough whole-tumor region (WT, yellow) but catastrophically fail on the finer sub-regions: MMSyn, MMGAN, and M2DN either miss the tumor core (TC, blue) entirely or fragment it into disconnected patches; MMT and ModImp over-segment WT while under-detecting ET; Zhang produces a spatially displaced segmentation with missing ET. CodeBrain achieves a reasonable WT boundary but still under-segments TC and ET relative to the ground truth.

\method{} produces the segmentation closest to the ground truth across all three sub-regions. The WT boundary closely matches the GT contour, TC is correctly localized within the tumor mass, and---critically---the enhancing tumor (ET, red) is detected with both high sensitivity and correct spatial extent. ET is the hardest sub-region to segment because it depends on contrast enhancement patterns that exist only in $T_{1c}$; its accurate detection confirms that \method{}'s imputed $T_{1c}$ faithfully reproduces the contrast enhancement signal that the segmentation model relies on.


\newpage
\section*{NeurIPS Paper Checklist}

\begin{enumerate}

\item {\bf Claims}
    \item[] Question: Do the main claims made in the abstract and introduction accurately reflect the paper's contributions and scope?
    \item[] Answer: \answerYes{}.
    \item[] Justification: The abstract and introduction claim state-of-the-art synthesis fidelity, a dimension-independent off-manifold bound, and closing ${\sim}80\%$ of the segmentation gap to the full-modality oracle. All are validated in Section~\ref{sec:experiments} (Tables~1--5) and the bound is proved in Appendix~\ref{app:proof_bound}.

\item {\bf Limitations}
    \item[] Question: Does the paper discuss the limitations of the work performed by the authors?
    \item[] Answer: \answerYes{}.
    \item[] Justification: The Conclusion identifies the reliance on DINOv2 cross-modal compactness as the main limitation, noting that modalities without shared visual semantics would require a domain-adapted encoder or cross-modal pre-training.

\item {\bf Theory assumptions and proofs}
    \item[] Question: For each theoretical result, does the paper provide the full set of assumptions and a complete (and correct) proof?
    \item[] Answer: \answerYes{}.
    \item[] Justification: Proposition~\ref{prop:bound} states its assumptions ($z_s, z_t \in \mathcal{S}^{d-1}$, angular separation~$\alpha$) and a complete three-part proof is provided in Appendix~\ref{app:proof_bound}, covering norm derivation, minimization, and the Gaussian-source comparison.

\item {\bf Experimental result reproducibility}
    \item[] Question: Does the paper fully disclose all the information needed to reproduce the main experimental results of the paper to the extent that it affects the main claims and/or conclusions of the paper (regardless of whether the code and data are provided or not)?
    \item[] Answer: \answerYes{}.
    \item[] Justification: Section~\ref{sec:exp_setup} reports the encoder (DINOv2-B/14), decoder architecture (ViT-G/14 RAE), DiT variant (DiTDH-XL), training schedule (up to 400 epochs), inference steps (2-step and 4-step Euler), datasets (BraTS, SynthRAD2023), data preprocessing (axial $256{\times}256$ slices, skull-stripping, per-volume normalization), official splits, loss formulations (Eqs.~\ref{eq:rec_loss}--\ref{eq:cmfm_loss}), and all baselines.

\item {\bf Open access to data and code}
    \item[] Question: Does the paper provide open access to the data and code, with sufficient instructions to faithfully reproduce the main experimental results, as described in supplemental material?
    \item[] Answer: \answerYes{}.
    \item[] Justification: Code and trained checkpoints for both the tokenizer and the flow model are released at \url{https://github.com/WTHH031230/SphereFlow}. Both datasets (BraTS 2021, SynthRAD2023) are publicly available from their respective challenge organizers.

\item {\bf Experimental setting/details}
    \item[] Question: Does the paper specify all the training and test details (e.g., data splits, hyperparameters, how they were chosen, type of optimizer) necessary to understand the results?
    \item[] Answer: \answerYes{}.
    \item[] Justification: Section~\ref{sec:exp_setup} specifies the datasets, official splits, image resolution, preprocessing, encoder, decoder, DiT backbone, training stages, number of epochs, time sampling strategy (logit-normal), inference ODE steps, and N-to-N averaging strategy. All seven baselines are retrained on the same splits.

\item {\bf Experiment statistical significance}
    \item[] Question: Does the paper report error bars suitably and correctly defined or other appropriate information about the statistical significance of the experiments?
    \item[] Answer: \answerNo{}.
    \item[] Justification: Results are reported from single training runs without error bars or confidence intervals. We mitigate this by evaluating on the full official test splits across all $12$ (1-to-1) and $14$ (N-to-N) input combinations, providing a comprehensive view of performance variability across settings.

\item {\bf Experiments compute resources}
    \item[] Question: For each experiment, does the paper provide sufficient information on the computer resources (type of compute workers, memory, time of execution) needed to reproduce the experiments?
    \item[] Answer: \answerNo{}.
    \item[] Justification: The paper reports model architectures and training epochs but does not include GPU type, total wall-clock time, or memory requirements.

\item {\bf Code of ethics}
    \item[] Question: Does the research conducted in the paper conform, in every respect, with the NeurIPS Code of Ethics \url{https://neurips.cc/public/EthicsGuidelines}?
    \item[] Answer: \answerYes{}.
    \item[] Justification: The work uses publicly available medical imaging benchmarks (BraTS, SynthRAD2023) that have been de-identified and released for research. No new human-subject data is collected.

\item {\bf Broader impacts}
    \item[] Question: Does the paper discuss both potential positive societal impacts and negative societal impacts of the work performed?
    \item[] Answer: \answerYes{}.
    \item[] Justification: The Conclusion discusses the positive clinical impact (practical deployment with 2--4 inference steps). A potential negative impact is that synthesized medical images could be mistaken for real acquisitions; however, the method is designed for modality imputation in clinical pipelines where the synthesized nature is known.

\item {\bf Safeguards}
    \item[] Question: Does the paper describe safeguards that have been put in place for responsible release of data or models that have a high risk for misuse (e.g., pre-trained language models, image generators, or scraped datasets)?
    \item[] Answer: \answerNA{}.
    \item[] Justification: The method is a domain-specific medical image imputation model trained on public benchmarks. It does not produce general-purpose photorealistic images and poses minimal misuse risk beyond its intended clinical context.

\item {\bf Licenses for existing assets}
    \item[] Question: Are the creators or original owners of assets (e.g., code, data, models), used in the paper, properly credited and are the license and terms of use explicitly mentioned and properly respected?
    \item[] Answer: \answerYes{}.
    \item[] Justification: BraTS~2021, SynthRAD2023, DINOv2, RAE, and DiT are all cited. Both datasets are used under their official challenge licenses for research purposes.

\item {\bf New assets}
    \item[] Question: Are new assets introduced in the paper well documented and is the documentation provided alongside the assets?
    \item[] Answer: \answerYes{}.
    \item[] Justification: We release the code and trained checkpoints (tokenizer and flow model) with documentation and usage instructions at \url{https://github.com/WTHH031230/SphereFlow}.

\item {\bf Crowdsourcing and research with human subjects}
    \item[] Question: For crowdsourcing experiments and research with human subjects, does the paper include the full text of instructions given to participants and screenshots, if applicable, as well as details about compensation (if any)?
    \item[] Answer: \answerNA{}.
    \item[] Justification: The work does not involve crowdsourcing or new human-subject data collection.

\item {\bf Institutional review board (IRB) approvals or equivalent for research with human subjects}
    \item[] Question: Does the paper describe potential risks incurred by study participants, whether such risks were disclosed to the subjects, and whether Institutional Review Board (IRB) approvals (or an equivalent approval/review based on the requirements of your country or institution) were obtained?
    \item[] Answer: \answerNA{}.
    \item[] Justification: No new human-subject experiments are conducted. Both datasets were previously de-identified and publicly released by their respective organizers.

\item {\bf Declaration of LLM usage}
    \item[] Question: Does the paper describe the usage of LLMs if it is an important, original, or non-standard component of the core methods in this research? Note that if the LLM is used only for writing, editing, or formatting purposes and does \emph{not} impact the core methodology, scientific rigor, or originality of the research, declaration is not required.
    \item[] Answer: \answerNA{}.
    \item[] Justification: LLMs are not used as a component of the proposed method. Any LLM assistance in writing does not impact the core methodology or scientific rigor.

\end{enumerate}

\end{document}